\documentclass[runningheads]{llncs}

\usepackage{eccv}

\usepackage{eccvabbrv}

\usepackage{graphicx}
\usepackage{booktabs}

\usepackage[accsupp]{axessibility}  

\usepackage{hyperref}
\usepackage{booktabs}   
\usepackage{multirow}
\usepackage{orcidlink}
\usepackage[table]{xcolor}
\usepackage{multirow}
\usepackage{graphicx}
\usepackage{tcolorbox}
\tcbuselibrary{skins,breakable}
\usepackage[accsupp]{axessibility}  
\newlength{\promptboxwidth}
\usepackage{xcolor}
\usepackage{tikz}
\usepackage{listings}

\lstdefinestyle{promptstyle}{
  basicstyle=\ttfamily\tiny,
  breaklines=true,
  breakatwhitespace=false,
  columns=fullflexible,
  keepspaces=true,
  showstringspaces=false,
  frame=single,
  framerule=0.4pt,
  framesep=2pt,
  rulecolor=\color{gray!65},
  backgroundcolor=\color{gray!3},
  linewidth=\promptboxwidth,
  xleftmargin=0pt,
  xrightmargin=0pt,
  aboveskip=0pt,
  belowskip=0pt
}

\newcommand{\prompttitle}[1]{%
  \noindent
  \begin{tikzpicture}
    \path[use as bounding box] (0,0) rectangle (\promptboxwidth,0.42cm);
    \fill[gray!65, rounded corners=2pt] (0,0) rectangle (\promptboxwidth,0.42cm);
    \node[
      text=white,
      font=\bfseries\footnotesize,
      align=center
    ] at ({\dimexpr\promptboxwidth/2\relax},0.21cm) {#1};
  \end{tikzpicture}%
  \par\vspace{-0.2em}
}

\begin{document}

\title{RubricRL: Simple Generalizable Rewards for Text-to-Image Generation} 

\titlerunning{RubricRL}

\author{
Xuelu Feng$^{1}$ \quad
Yunsheng Li$^{2}$ \quad
Ziyu Wan$^{2}$ \quad
Zixuan Gao$^{3}$ \quad
Junsong Yuan$^{1}$ \\
Dongdong Chen$^{2}$$^*$ \quad
Chunming Qiao$^{1}$ \\
}
\authorrunning{X. Feng et al.}

\institute{{$^1$University at Buffalo} \quad  {$^2$Microsoft AI} \quad {$^3$Nikola Tesla STEM High School} \\
{\small $^*$Corresponding Author} \\
\email{\{xuelufen,jsyuan,qiao\}@buffalo.edu}, 
 {\{yunshengli, ziyuwan,dongdong.chen\}@microsoft.com}, zixuan.go.1@gmail.com} 

\maketitle
\vspace{-2em}
\begin{figure}
\centering
\includegraphics[width=\linewidth]{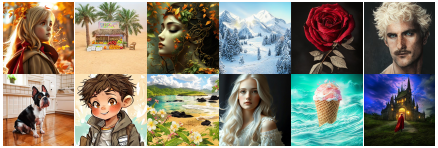}
\captionof{figure}{Visual examples of our RubricRL on two language backbones. Equipped with interpretable and user-controlled criteria, RubricRL improves SFT models' performance to generate high-quality images.}
\label{fig:teaser}
\end{figure}
\vspace{-3em}
\begin{abstract}


Reinforcement learning (RL) has recently emerged as a promising approach for aligning text-to-image generative models with human preferences. A key challenge, however, lies in designing effective and interpretable rewards. Existing methods often rely on either composite metrics (\textit{e.g.}, CLIP, OCR, and realism scores) with fixed weights or a single scalar reward distilled from human preference models, which can limit interpretability and flexibility. We propose RubricRL, a simple and general framework for rubric-based reward design that offers greater interpretability, composability, and user control. Instead of using a black-box scalar signal, RubricRL dynamically constructs a structured rubric for each prompt—a decomposable checklist of fine-grained visual criteria such as object correctness, attribute accuracy, OCR fidelity, and realism—tailored to the input text. Each criterion is independently evaluated by a multimodal judge (\textit{e.g.}, o4-mini), and a prompt-adaptive weighting mechanism emphasizes the most relevant dimensions. This design not only produces interpretable and modular supervision signals for policy optimization (\textit{e.g.}, GRPO or DiffusionNFT), but also enables users to directly adjust which aspects to reward or penalize. Experiments with autoregressive and diffusion text-to-image models demonstrate that RubricRL improves prompt faithfulness, visual detail, and generalizability, while offering a flexible and extensible foundation for interpretable RL alignment across text-to-image architectures.
\keywords{Rubric reward \and Reinforcement learning \and Text-to-image generation}

\end{abstract}    

\section{Introduction}
\label{sec:intro}

Reinforcement learning (RL) has recently emerged as a promising approach for aligning generative models~\cite{sohl2015deep,ho2020denoising,song2019generative,song2020score,rombach2022high,avrahami2022blended,choi2021ilvr, chen2018pixelsnail,chen2025janus,ramesh2021zero,sun2024autoregressive,tian2024visual} with human preferences. In large language models, frameworks such as RLHF~\cite{ouyang2022training} and RLVF~\cite{yu2025dapo,shao2024deepseekmath} have demonstrated that policy optimization guided by preference-based feedback can significantly enhance faithfulness, style, and usability. Extending this paradigm to text-to-image generation, including both autoregressive (AR) and diffusion architectures, offers a principled way to optimize models directly for human-aligned visual quality rather than likelihood-based objectives. However, the effectiveness of RL in visual domains critically depends on reward design: constructing evaluation signals that are accurate, interpretable, and generalizable across prompts, domains, and architectures remains a core challenge.

\begin{figure}[t]
\centering
\includegraphics[width=\linewidth]{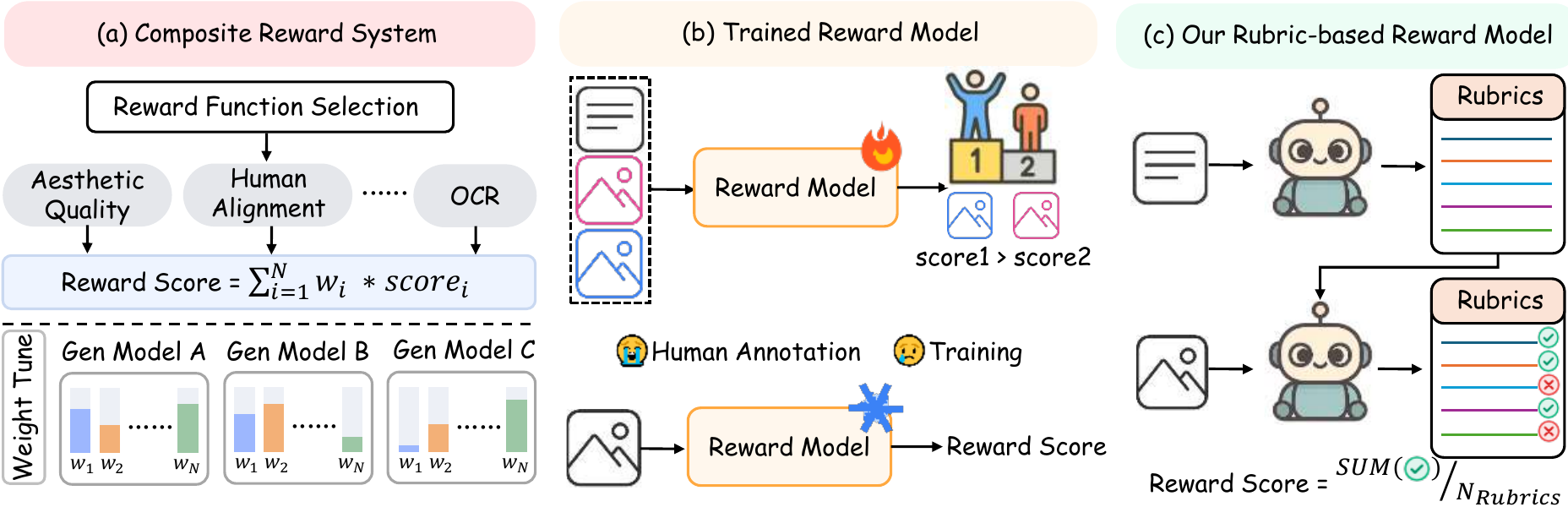}
\caption{Comparison of RubricRL with prior autoregressive (AR) reward formulations. (a) Multi-reward pipelines combine CLIP, OCR, alignment and realism metrics but require fragile weight tuning and often miss fine-grained attributes. (b) Unified scalar models collapse diverse objectives into a single learned score, simplifying optimization but reducing interpretability and adaptability. (c) RubricRL replaces both with a decomposable, prompt-adaptive rubric—an explicit checklist of visual criteria (counting, attributes, OCR/text fidelity, realism). Each criterion is scored independently and integrated into RL framework to provide interpretable, variance-aware supervision that improves detail, prompt faithfulness, and debuggability.}
\label{fig:comparison}
\end{figure}

Existing text-to-image RL frameworks can be broadly categorized into multi-reward mixtures and unified scalar reward models. As shown in Fig.~\ref{fig:comparison} (a), multi-reward systems (\textit{e.g.}, X-Omni~\cite{geng2025x}, AR-GRPO~\cite{yuan2025ar}, Flow-GRPO~\cite{liu2025flow}, DanceGRPO~\cite{xue2025dancegrpo}) combine heterogeneous objectives, such as CLIP-based image–text similarity~\cite{radford2021learning}, OCR accuracy~\cite{cui2025paddleocr}, realism~\cite{wu2023human}, and attribute consistency, to jointly encourage alignment and visual quality. While such approaches improve coverage, they depend on manually tuned weighting schemes that can be brittle across prompts and domains, and offer limited interpretability. Unified reward models (\textit{e.g.}, OneReward~\cite{gong2025onereward}, Pref-GRPO~\cite{wang2025pref}, LLaVA-Reward~\cite{zhou2025multimodal}) in Fig.~\ref{fig:comparison} (b), instead learn a single scalar reward from pairwise human preference data. This simplifies optimization but can obscure the reasoning behind rewards, limit extensibility, and make it difficult for developers to control which visual aspects are prioritized.

In this paper, we propose RubricRL, a simple and general framework for rubric-based rewards design in text-to-image models as illustrated in Fig.~\ref{fig:comparison} (c). Rather than relying on opaque scalar signals, RubricRL dynamically selects a structured rubric for each prompt, \textit{i.e.}, a decomposable checklist of fine-grained visual criteria such as object correctness, attribute accuracy, OCR fidelity, compositional coherence, and realism. Each criterion is independently evaluated by a multimodal judge (\textit{e.g.}, GPT-o4-mini), while a prompt-adaptive weighting mechanism highlights the most relevant dimensions. This produces interpretable, modular supervision signals that integrate naturally into policy optimization frameworks such as GRPO~\cite{guo2025deepseek}, PPO~\cite{schulman2017proximal} or DiffusionNFT~\cite{zheng2025diffusionnft}.

By expressing rewards in human-readable and decomposable form, RubricRL transforms reward evaluation from a black-box heuristic into an auditable process, where developers can directly inspect, extend, or adjust which aspects of generation are rewarded or penalized. The rubric structure also facilitates per-criterion diagnostics, providing transparency into model behavior and simplifying both evaluation and debugging.

RubricRL is architecture-agnostic and compatible with both diffusion and autoregressive text-to-image models. The rubric outputs further support variance-aware group advantages, leading to robust updates even under long-horizon rollouts. Its prompt-adaptive design ensures that each reward vector reflects the salient aspects of the input text, such as numerals, named entities, styles, or embedded text, without requiring manual tuning.

We validate this simple yet effective idea on an autoregressive (AR) text-to-image model, and further demonstrate its generalizability on diffusion-based generators. Experiments show that RubricRL improves prompt faithfulness, compositional accuracy, and visual realism, while maintaining high generalizability across architectures and datasets. Compared to prior multi-reward or unified-reward approaches, RubricRL achieves more consistent optimization behavior and enables controllable, interpretable reward shaping. Fig.~\ref{fig:teaser} provides visualization samples of our method, illustrating high visual quality.

In summary, RubricRL contributes:
\begin{itemize}
\item A generalizable rubric-based reward design applicable to both AR and diffusion text-to-image models;
\item A prompt-adaptive, decomposable supervision framework that enhances interpretability and composability;
\item A developer-controllable and auditable framework that exposes rubric categories as prompt-level controls, allowing users to tailor reward dimensions via simple system-prompt edits.
\end{itemize}

By operationalizing alignment through dynamically generated rubrics of explicit visual criteria, RubricRL makes reinforcement learning for text-to-image generation more interpretable, extensible, and developer-guided, offering a unified foundation for aligning visual generation with human intent.

\section{Related work}
\label{sec:related_work}
\subsection{Text-to-Image Generation Methods.} 
Text-to-image (T2I) generation has seen significant progress through both diffusion based \cite{nichol2021glide,rombach2022high,gu2022vector,chang2023muse,wei2024enhancing,wu2025qwen} and autoregressive (AR) architectures \cite{yu2022scaling,dong2023dreamllm,zhou2024transfusion,wu2025janus}. Diffusion models iteratively refine latent representations conditioned on text prompts, achieving high-quality and photorealistic images. Variants such as Stable Diffusion \cite{rombach2022high} and flow-based extensions \cite{esser2024scaling,li2025omniflow,flux2024} provide diverse styles, controllable generation, and strong fidelity at both global and local levels. Autoregressive approaches, on the other hand, represent images as sequences of discrete tokens and model the joint distribution of text and image tokens using a single transformer backbone. Early hybrid designs, such as DreamLLM~\cite{dong2023dreamllm}, paired AR text encoders with separate diffusion decoders. More recent unified AR models, including Chameleon~\cite{karypis1999chameleon}, Emu3~\cite{wang2024emu3}, TransFusion~\cite{zhou2024transfusion}, and Janus~\cite{wu2025janus}, integrate visual tokenization and autoregressive modeling in one architecture. These models allow direct mapping between text tokens and visual outputs, enabling flexible control and fine-grained generation. In this paper, we propose a novel reward design for reinforcement learning in text-to-image models, and demonstrate their effectiveness using both unified AR and diffusion text-to-image models.

\subsection{Reinforcement Learning for Text-to-Image Generation.}
Maximum-likelihood training often under-optimizes user-salient qualities, such as semantic faithfulness, compositional accuracy, and aesthetics. Reinforcement learning (RL) offers task-aligned feedback that directly optimizes for human-relevant properties beyond likelihood. In diffusion-based text-to-image models, RL methods, such as Flow-GRPO~\cite{liu2025flow}, DanceGRPO~\cite{xue2025dancegrpo}, and DifusionNFT~\cite{zheng2025diffusionnft}, have improved alignment by fine-tuning the velocity with preference or metric-based rewards. Recently, RL has also been applied to unified AR T2I models \cite{wang2025simplear}, where policy gradients act directly on next-token probabilities, enabling end-to-end credit assignment and fine-grained control over generated images. 

The design of the reward function is central to effective reinforcement learning in text-to-image models. One line of work aggregates heterogeneous signals—such as CLIP-based image–text alignment~\cite{radford2021learning}, OCR/text correctness~\cite{wei2024general,cui2025paddleocr}, multimodal VLM judges (\textit{e.g.}, Qwen2.5-VL-32B~\cite{bai2025qwen2}), aesthetic and realism metrics~\cite{yang2022maniqa}, and human-preference surrogates~\cite{wu2023human}. While comprehensive, these multi-reward mixtures demand careful weighting and tuning, which can destabilize optimization and obscure per-aspect failures. Another direction trains unified preference models~\cite{wang2025unified,wang2025pref,zhou2025multimodal} to predict a single scalar human-aligned score from paired image outputs, simplifying optimization but relying on costly human annotations and limited scalability. In this work, we propose a rubric-based reward that is simple, generalizable, and interpretable. For each prompt, a compact rubric defines aspect-wise criteria—such as text alignment/OCR accuracy, object count, spatial relations, and overall coherence/quality. Each criterion is scored independently by a dedicated evaluator, and a transparent aggregation produces the final reward. This design is more prompt-adaptive, decomposable, and interpretable, while providing user-controllable and auditable feedback. While several concurrent works \cite{gunjal2025rubrics,huang2025reinforcement} investigate rubric-based rewards in natural language processing, to the best of our knowledge, we are the first to propose rubric based rewards in text-to-image RL.
\section{Method}
\begin{figure}[t]
  \centering
  \includegraphics[width=\linewidth]{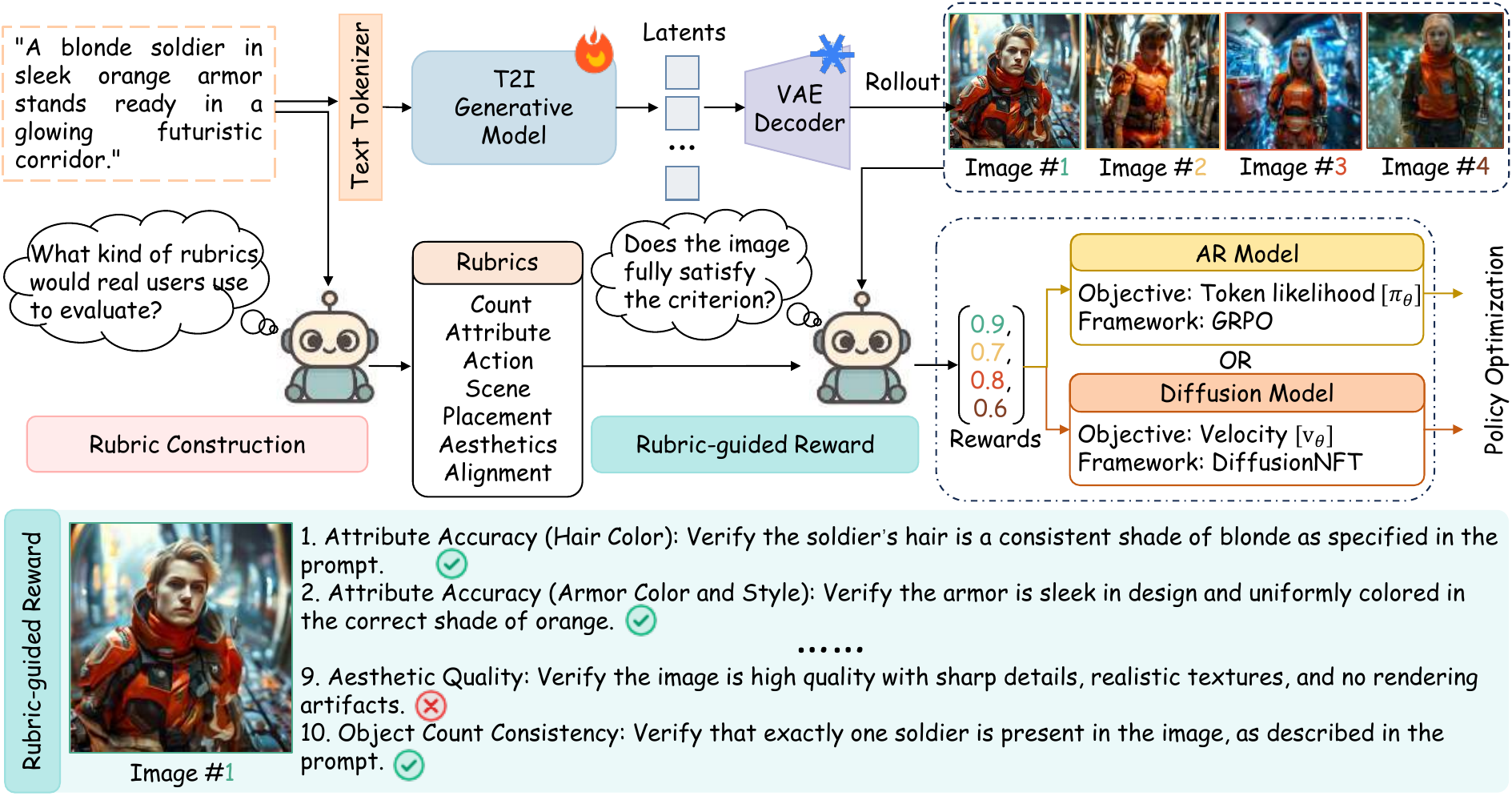}
  \caption{Overview of the proposed method. We propose a simple, general rubric generation pipeline and rubric-based reward model for text-to-image generation.}
  \label{fig:pipeline}
  \vspace{-1.em}
\end{figure}

In this section, we present RubricRL as a general post-training framework for text-to-image generation that applies to both autoregressive (AR) and diffusion-based backbones (Fig.~\ref{fig:pipeline}). The core components---rubric construction and rubric-guided reward---are shared across model families, while the RL optimizer differs (GRPO for AR and DiffusionNFT for diffusion). For clarity, we describe the method using an AR instantiation in the main text and defer diffusion-specific implementation details to the supplementary material.

\subsection{Overall Architecture}

As illustrated in Fig.~\ref{fig:pipeline}, given an input text prompt $p$, we first tokenize it into a sequence of text tokens and feed them into a backbone text-to-image generator $\pi_\theta$ (either an AR token generator or a diffusion model) to produce a generated image $I$. In this paper, we primarily focus on post-RL fine-tuning of $\pi_\theta$ to further enhance its output quality, where designing an effective, reliable, and interpretable reward function is the key challenge. Existing methods typically employ one or multiple specialized models to evaluate different aspects of image quality, such as CLIP-based image--text semantic alignment reward~\cite{radford2021learning} ($R_\text{clip}(I, p)$), OCR accuracy~\cite{cui2025paddleocr} ($R_\text{ocr}(I, p)$), and realism~\cite{wu2023human}. However, this approach has notable drawbacks: (1) deploying multiple specialized models is computationally expensive and difficult to scale to additional aspects; (2) it requires careful reward calibration and reweighting. Recent works have attempted to learn a single reward model from pairwise human preference data, simplifying optimization but offering limited extensibility due to high annotation costs and poor interpretability.


Motivated by the strong multimodal understanding capabilities of modern vision--language models (VLMs), we propose a simple and unified \emph{rubric-based reward model}, denoted $R_\text{rubric}(I, p, \mathcal{C}(p))$. This model replaces the ensemble of task-specific evaluators with a single reasoning-capable grader. Rather than relying on fixed sub-models, our approach automatically constructs a set of interpretable, prompt-adaptive criteria termed \emph{``rubrics''} that capture the essential quality requirements for each prompt $p$. In detail, given a prompt $p$, a \emph{Rubric Generation Model} $\mathcal{G}$ generates a set of evaluation rubrics $\mathcal{C}(p)=\mathcal{G}(p)$, where $\mathcal{C}(p)=\{c_1,c_2,\dots,c_M\}$ defines $M$ prompt-specific criteria spanning dimensions such as object count, attribute accuracy, text/OCR fidelity, spatial relations, aesthetics, and style consistency.

In reinforcement learning (RL), the objective is to adjust the model parameters $\theta$ to maximize the expected rubric reward over the distribution of prompts:
\begin{equation}
\max_\theta \ \mathbb{E}_{p \sim \mathcal{D},\, I \sim \pi_\theta(\cdot|p)} \big[\, R_\text{rubric}(I, p, \mathcal{C}(p)) \,\big],
\end{equation}
where $\mathcal{D}$ denotes the set of prompts and a \emph{rollout} corresponds to one sampled image from $\pi_\theta$ given $p$. Compared to composite reward systems, our rubric-based formulation offers three key advantages:
(1) \textbf{Simplicity}: it eliminates the need for multiple task-specific graders;  
(2) \textbf{Adaptivity}: rubrics are dynamically generated for each prompt, ensuring relevance to diverse intents; and  
(3) \textbf{Interpretability}: each reward component aligns with a human-readable evaluation criterion, enabling transparent model diagnostics and controllable optimization.



\subsection{Rubric-based Reward}
\label{sec:rubric-generation}

The rubric based reward function proceeds in two stages. First, a \emph{Rubric Generation Model} \(\mathcal{G}\) interprets the user prompt \(p\) and produces a set of candidate evaluation rubrics \(\mathcal{C}(p)\). Second, a multimodal LLM grader implements the \emph{Rubric-Based Reward} \(R_{\text{rubric}}(I,p,\mathcal{C}(p))\) that scores a generated image \(I\) against each rubric in \(\mathcal{C}(p)\). In this paper, we employ GPT-o4-mini to fulfill both roles, generating prompt-specific rubrics and providing per-criterion judgments that are aggregated into a scalar reward.


\noindent\textbf{Rubric construction}. Given a user prompt $p$, we ask GPT-o4-mini to generate a list of rubrics. Each rubric entry contains a short eval key that targets a specific aspect (\textit{e.g.}, OCR alignment, object count, spatial relations, aesthetics) and a concise description of what to check in the image. 

To promote diversity and reduce positional bias during rubric generation, we randomly permute the evaluation aspects in the rubric generation prompt and query GPT-o4-mini multiple times. In each round, the model produces a set of rubrics (we request 10 per query; because a prompt may describe multiple objects or attributes, the model may output multiple rubrics for one eval key to ensure adequate coverage). We aggregate all valid key–criterion pairs across runs into a unified rubric pool, discarding ambiguous or malformed entries. Finally, to remove redundancy and focus on the most important signals, we ask GPT-o4-mini to choose the top-10 most relevant and critical criteria for evaluating images generated from the user prompt $p$. 

\noindent\textbf{Rubric-guided reward.} Given a generated image $I$, its corresponding text prompt $p$ and the rubric pool $\mathcal{C}$, we simply ask GPT-o4-mini again to output a single score $y_i \in \{0, 1\}$ for each criterion to reflect whether the generated image fully satisfies this rubric ($y_i=1$) or not ($y_i=0$). The overall rubric reward is computed as the normalized mean of:
\begin{equation}
R(I, p, \mathcal{C}) = \frac{1}{M}\sum_{i=1}^{M} y_i, \quad M=10
\end{equation}

\subsection{Reinforcement Learning with GRPO}

To align the autoregressive image generator with rubric-based rewards, we employ \textbf{Group Relative Policy Optimization (GRPO)}~\cite{shao2024deepseekmath}, a variant of PPO designed for stable optimization over grouped rollouts. For diffusion-based models, we adopt \textbf{DiffusionNFT}~\cite{zheng2025diffusionnft} as the RL strategy; details are provided in the supplementary material. For each prompt, the set of generated rollouts forms a group, and the reward of each rollout is normalized relative to the group to reduce variance and improve credit assignment. Concretely, let $\pi_\theta$ denote the current policy and $R_i$ the rubric reward for rollout $i$ in group $g$. GRPO computes the relative advantage
\begin{equation}
\begin{aligned}
A_i
&= \frac{R_i - \bar{R}_g}
{\sqrt{\frac{1}{|g|-1}\sum_{j\in g}\left(R_j - \bar{R}_g\right)^2}}, \
\bar{R}_g= \frac{1}{|g|}\sum_{k\in g} R_k,
\end{aligned}
\label{eq:adv}
\end{equation}
and updates the policy by maximizing a clipped objective similar to PPO:
\begin{equation}
\mathcal{L}(\theta) = \mathbb{E}_i \Big[ \min\big( r_i(\theta) A_i, \, \mathrm{clip}(r_i(\theta), 1-\epsilon, 1+\epsilon) A_i \big) \Big],
\label{eq:grpo_loss}
\end{equation}
where $r_i(\theta) = \frac{\pi_\theta(a_i|s_i)}{\pi_{\theta_\mathrm{old}}(a_i|s_i)}$, $a_i$ and $s_i$ are the sampled action and state corresponding to rollout $i$, and $\epsilon$ is the PPO clipping parameter. By leveraging this group-relative advantage, GRPO stabilizes training across prompts, making the model robust to heterogeneous reward scales and noisy evaluations. Combined with our rubric-based reward and dynamic rollout selection strategy described below, we find GRPO can effectively guide the generative model toward images that are both human-aligned and high-quality.

\subsection{Dynamic Rollout Sampling}
\label{sec:dyn-select}

As discussed above, the target policy model $\pi_\theta$ in GRPO explores the generation space by sampling multiple rollouts, each yielding a reward $R_i$ used for advantage computation. In the original GRPO design, all $N$ rollouts from a single prompt are grouped together for policy updates, \textit{i.e.}, $|g| = N$. Subsequent works introduce over-sampling and filtering strategies to improve training efficiency. For instance, DAPO~\cite{yu2025dapo} adopts a \emph{prompt-level} over-sampling approach: it generates $N$ rollouts per prompt and discards prompts whose rollouts all have accuracy $1$ or $0$, thereby retaining only moderately difficult prompts for policy optimization. Formally, DAPO selectively sample prompts used in training while still using all rollouts from each retained prompt for RL updates.

In this paper, we propose a new \emph{rollout-level} dynamic sampling mechanism, where selection occurs within the rollouts of a single prompt rather than filtering entire prompts. Specifically, given a text prompt, instead of sampling only $N$ rollouts, we oversample $N'$ rollouts ($N' > N$) and selectively use a subset of $N$ representative rollouts for policy updates. To balance quality and diversity, we adopt a hybrid selection strategy: we take the top-$K$ high-reward rollouts and randomly sample the remaining $N-K$ rollouts from the others to encourage diversity. Formally, the rollout group $g$ is constructed as
\begin{equation}
g \;=\; \{\tau_{(1)}, \dots, \tau_{(K)}\} \;\cup\; \mathrm{RS}\big(\{\tau_{(K+1)}, \dots, \tau_{(N')}\}, N-K\big),
\end{equation}
where $\mathrm{RS}$ denotes random sampling. Empirically, we observe this hybrid design achieves a better balance between stability and diversity, achieving better model quality. As a result, the loss in Eq.~\ref{eq:grpo_loss} is computed over a more representative and informative subset of rollouts, leading to more consistent and efficient learning compared to both the original GRPO and the prompt-level filtering in DAPO.

\section{Experiments}
\subsection{Implementation Details}

Following SimpleAR~\cite{wang2025simplear}, we construct a training set of 11,000 images sampled from JourneyDB~\cite{sun2023journeydb} and Synthetic-1M~\cite{Egan_Dalle3_1_Million_2024}. We re-caption all images with GPT-o4-mini to obtain diverse prompt lengths, and randomly sample one caption per image during training. All experiments are conducted on $8$ NVIDIA A100 GPUs.

\noindent\textbf{Autoregressive (AR) backbones.}
We evaluate two AR backbones: Phi3-3.8B~\cite{abdin2024phi} and Qwen2.5-0.5B~\cite{team2024qwen25}, both initialized from SFT checkpoints. For image decoding, we use LlamaGen's VQ decoder~\cite{sun2024autoregressive} for Phi3-3.8B and Cosmos-Tokenizer~\cite{agarwal2025cosmos} for Qwen2.5-0.5B. The output resolutions are $512$ and $1024$, respectively. RL training uses the TRL framework~\cite{vonwerra2022trl} with learning rate $1\mathrm{e}{-5}$, warm-up ratio $0.1$, batch size $28$, and $3$ epochs. We use dynamic rollout sampling by selecting $4$ candidates from $16$ rollouts per prompt. During inference, we apply classifier-free guidance (CFG)~\cite{ho2022classifier} with scale 7.5.

\noindent\textbf{Diffusion backbones.}
For diffusion-based text-to-image models, we additionally fine-tune Qwen-Image~\cite{wu2025qwen} and FLUX-1.0 [dev]~\cite{blackforestlabs2024flux1dev} using DiffusionNFT~\cite{zheng2025diffusionnft}. We adopt LoRA with rank $r{=}32$ and scaling factor $\alpha{=}64$. Each epoch consists of $48$ groups with group size $24$. We use $12$ rollout sampling steps, and $40$ denoising steps. We set the learning rate to $3\mathrm{e}{-4}$ and batch size to $3$. In inference stage, CFG is applied to Qwen-Image with a scale of 7.5, and we use the default guidance scale of 3.5 for FLUX.

\begin{table}[t]
\caption{Evaluation of text-to-image generation trained on Phi3 (3.8B) and Qwen2.5 (0.5B) as AR backbone on the GenEval.}
\vspace{-0.5em}
\centering
\scriptsize
\setlength{\tabcolsep}{1pt}
\begin{tabular}{lccccccc}
\toprule
Method & Single Obj. & Two Obj. & Counting & Colors & Position & Color Attr. & Overall \\
\midrule
Phi3 (3.8B)~\cite{abdin2024phi} & 0.9938 & 0.8939 & 0.4562 & 0.9255 & 0.7200 & 0.5850 & 0.7624 \\
+ CLIPScore~\cite{radford2021learning} & 0.9938 & 0.9242 & 0.5250 & 0.9362 & 0.7725 & 0.7000 & 0.8086 \\
+ HPSv2~\cite{wu2023human}  & 0.9906 & 0.8813 & 0.5125 & 0.9441 & 0.7675 & 0.7205 & 0.8035 \\
+ Unified Reward~\cite{wang2025unified}  & 0.9969 & 0.9318 & 0.4156 & 0.9388 & 0.8150 & 0.7000 & 0.7997 \\
+ LLaVA-Reward-Phi~\cite{zhou2025multimodal}  & 0.9844 & 0.8864 & 0.4719 & 0.9176 & 0.7250 & 0.5975 & 0.7638 \\
+ AR-GRPO~\cite{yuan2025ar}  & 0.9938 & 0.8712 & 0.5406 & \textbf{0.9574} & 0.8075 & 0.6300 & 0.8001 \\
+ X-Omni~\cite{geng2025x}  & 0.9969 & 0.9192 & 0.4719 & 0.9548 & 0.8175 & 0.6875 & 0.8080 \\
\rowcolor{gray!20}
+ RubricRL & \textbf{1.0000} & \textbf{0.9343} & \textbf{0.6125} & 0.9415 & \textbf{0.8275} & \textbf{0.7650} & \textbf{0.8468} \\
\midrule
Qwen2.5 (0.5B)~\cite{team2024qwen25} & 0.9625 & 0.5303 & 0.2500 & 0.7606 & 0.3575 & 0.2825 & 0.5239 \\
+ CLIPScore~\cite{radford2021learning} & 0.9656 & 0.6162 & 0.2750 & \textbf{0.8404} & 0.3825 & 0.3250 & 0.5674 \\
+ HPSv2~\cite{wu2023human} & 0.9750 & 0.6465 & 0.2438 & 0.8005 & 0.3875 & 0.2825 & 0.5560 \\
+ Unified Reward~\cite{wang2025unified}  & 0.9625 & 0.6288 & 0.2656 & 0.8191 & 0.4050 & 0.3550 & 0.5727 \\
+ LLaVA-Reward-Phi~\cite{zhou2025multimodal}  & 0.9625 & 0.5303 & 0.2500 & 0.7606 & 0.3575 & 0.2825 & 0.5239 \\
+ AR-GRPO~\cite{yuan2025ar}  & 0.9656 & 0.5682 & \textbf{0.2969} & 0.8378 & 0.3825 & 0.3050 & 0.5593 \\
+ X-Omni~\cite{geng2025x}  & 0.9812 & 0.5960 & 0.2219 & 0.8085 & 0.4125 & 0.3200 & 0.5567 \\
\rowcolor{gray!20}
+ RubricRL & \textbf{0.9844} & \textbf{0.6616} & 0.2469 & 0.8378 & \textbf{0.4825} & \textbf{0.3950} & \textbf{0.6014} \\
\bottomrule
\end{tabular}
\label{tab:geneval}

\end{table}

\begin{table}[t]
\caption{Evaluation of text-to-image generation trained on Phi3 (3.8B) and Qwen2.5 (0.5B) as AR backbone on the DPG-Bench.}
\centering
\vspace{-0.5em}
\scriptsize
\setlength{\tabcolsep}{6pt}
\begin{tabular}{llcccccc}
\toprule
Method & Global & Entity & Attribute & Relation & Other & Overall \\
\midrule
Phi3 (3.8B)~\cite{abdin2024phi} & \textbf{84.80} & 87.90 & 88.18 & 93.30 & 82.00 & 81.25 \\
+ CLIPScore~\cite{radford2021learning}  & 81.76 & 89.95 & 89.42 & 93.50 & 86.00 & 84.15 \\
+ HPSv2~\cite{wu2023human}  & 82.98 & 90.71 & 89.94 & 93.19 & 87.60 & 84.85 \\
+ Unified Reward~\cite{wang2025unified}   & 82.37 & 89.94 & 89.50 & 93.93 & 85.20 & 84.06  \\
+ LLaVA-Reward-Phi~\cite{zhou2025multimodal}&  82.98 & 88.06 & 87.83 & 92.50 & 79.20 & 81.51  \\
+ AR-GRPO~\cite{yuan2025ar}  & 82.37 & 89.05 & 90.00 & 93.08 & \textbf{88.00} & 83.81 \\
+ X-Omni~\cite{geng2025x}  & 84.19 & 89.66 & 89.12 & 93.69 & 86.00 & 84.05 \\
\rowcolor{gray!20}
+ RubricRL (Ours)  & 83.28 & \textbf{91.88} & \textbf{90.07} & \textbf{94.73} & 85.20 & \textbf{86.07}
\\
\midrule
Qwen2.5 (0.5B)~\cite{team2024qwen25}  & 84.78 & 84.74 & 86.41 & 87.34 & 84.27 & 78.02 \\
+ CLIPScore~\cite{radford2021learning}  & 80.55 & 87.22 & 86.19 & 91.33 & 67.60 & 79.78  \\
+ HPSv2~\cite{wu2023human}  & 78.42 & 87.29 & 85.04 & 91.45 & 68.80 & 80.23 \\
+ Unified Reward~\cite{wang2025unified}   & 79.03 & 87.09 & 85.24 & 90.68 & 69.20 & 79.69 \\
+ LLaVA-Reward-Phi~\cite{zhou2025multimodal}& \textbf{84.78} & 84.74 & 86.41 & 87.34 & \textbf{84.27} & 78.02 \\
+ AR-GRPO~\cite{yuan2025ar}  & 80.24 & 86.75 & 85.95 & \textbf{92.02} & 69.35 & 79.74 \\
+ X-Omni~\cite{geng2025x}  & 79.33 & 87.03 & 85.34 & 91.72 & 72.40 & 79.92 \\
\rowcolor{gray!20}
+ RubricRL (Ours)  & 79.33 & \textbf{88.48} & \textbf{86.55} & 91.37 & 68.00 & \textbf{81.43} \\
\bottomrule
\end{tabular}

\label{tab:dpgbench}
\vspace{-1em}
\end{table}

\begin{table}[ht]
\centering
\caption{Comparison on diffusion-based generators. RubricRL consistently improves the diffusion-based Qwen-Image (4B) and FLUX (12B) baselines under the same training setting in DiffusionNFT.}
\vspace{-0.5em}
\scriptsize
\begin{tabular}{lccccccc}
\toprule
Method & Single Obj. & Two Obj. & Counting & Colors & Position & Color Attr. & Overall \\
\midrule
Qwen-Image (4B) ~\cite{wu2025qwen} & 1.0000 & 0.9242 & 0.7125 & 0.9521 & 0.7750 & 0.6800 & 0.8406 \\
+ CLIPScore & 0.9938 & 0.9394 & 0.7219 & 0.9628 & 0.7625 & 0.7325 & 0.8521 \\
+ HPSv2 & 0.9594 & 0.9293 & 0.6500 & 0.9122 & 0.6550 & 0.6500 & 0.7937 \\
+ UnifiedReward &0.9688 & 0.9217 & 0.7094 & 0.9229 & 0.6750 & 0.7025 & 0.8167 \\
+ LLaVA-Reward-Phi& 0.9969 & 0.9268 & 0.7344 & 0.9574 & 0.7250 & 0.7075 & 0.8405 \\
+ AR-GRPO & 0.9844 & 0.9369 & 0.7719 & 0.9441 & 0.7450 & 0.7275 & 0.8516 \\
+ X-Omni & 0.9938 & 0.9369 & 0.7438 & 0.9441 & 0.7725 & 0.7300 & 0.8535 \\
\rowcolor{gray!20}
+ RubricRL (Ours)  & \textbf{1.0000} & \textbf{0.9646} & \textbf{0.8562} & \textbf{0.9734} & \textbf{0.8275} & \textbf{0.8325} & \textbf{0.9091} \\
\midrule
FLUX.1 Dev (12B)~\cite{blackforestlabs2024flux1dev} & 0.9844 & 0.9293 & 0.7531 & 0.9255 & 0.6900 & 0.5900 & 0.8121 \\
+ CLIPScore & 0.9875 & 0.9697 & 0.8062 & 0.9521 & 0.7250 & 0.6700 & 0.8518 \\
+ HPSv2 & 0.9875 & 0.9444 & 0.6906 & 0.9096 & 0.6325 & 0.6525 & 0.8024 \\
+ UnifiedReward & 0.9906 & 0.9571 & 0.7688 & 0.9282 & 0.6700 & 0.6525 & 0.8279 \\
+ LLaVA-Reward-Phi & 0.9719 & 0.9697 & 0.7094 & 0.9362 & 0.6750 & 0.6850 & 0.8245 \\
+ AR-GRPO & 0.9750 & 0.9672 & 0.7188 & 0.9255 & 0.6850 & 0.7075 & 0.8298 \\
+ X-Omni & 0.9844 & 0.9495 & 0.7656 & 0.9335 & 0.7050 & 0.6625 & 0.8334 \\
\rowcolor{gray!20}
+ RubricRL (Ours)   &  \textbf{0.9969} & \textbf{0.9747} & \textbf{0.8438} & \textbf{0.9707} & \textbf{0.7800} & \textbf{0.7975} & \textbf{0.8969} \\
\bottomrule
\end{tabular}
\label{tab:diffusion_gen}
\vspace{-1em}
\end{table}

\begin{table}[ht]
\caption{Evaluation of text-to-image generation trained on Qwen-Image (4B) and FLUX.1 Dev (12B) as diffusion-based flow matching backbone on the DPG-Bench.}
\centering
\scriptsize

\setlength{\tabcolsep}{6pt}
\begin{tabular}{llcccccc}
\toprule
Method & Global & Entity & Attribute & Relation & Other & Overall \\
\midrule
Qwen-Image (4B)~\cite{wu2025qwen} 
& 78.72 & 86.47 & 88.68 & 93.00 & 81.60 & 82.40 \\
+ CLIPScore~\cite{radford2021learning} & 80.55 & 92.16 & \textbf{90.40}& 93.81 & 86.00 & 85.09 \\
+ HPSv2~\cite{wu2023human}  & 78.72 & 92.23 & 87.85 & 93.73 & 85.60 & 85.26 \\
+ Unified Reward~\cite{wang2025unified}  & 79.94 &  92.15& 90.30 & 93.93 & \textbf{86.40} & 85.96 \\
+ LLaVA-Reward-Phi~\cite{zhou2025multimodal} &82.07  & 89.31 & 88.88 & 92.69 & 86.00 & 81.29 \\
+ AR-GRPO~\cite{yuan2025ar} & 81.54 & 91.10 & 89.47 & 93.31 & 83.30 & 84.93 \\
+ X-Omni~\cite{geng2025x} & 83.13 & 91.83 & 89.80 & 93.97 & 84.30 & 85.80 \\
\rowcolor{gray!20}
+ RubricRL (Ours)  
& \textbf{82.37} & \textbf{92.63} & 89.88 &\textbf{94.82 } & 82.80 & \textbf{86.23}\\
\midrule
FLUX.1 Dev (12B)~\cite{blackforestlabs2024flux1dev}   & 81.46 &76.91  & 79.32 & 84.91 & 50.00 & 63.99 \\
+ CLIPScore~\cite{radford2021learning}   & 79.03 & 76.06 & 80.06 & 86.69 & 49.20 & 65.82 \\
+ HPSv2~\cite{wu2023human}  & 79.94 & \textbf{79.19} & \textbf{80.55} & 86.07 & 51.60 & 66.65  \\
+ Unified Reward~\cite{wang2025unified}    & 82.07 & 78.79 &80.33  & 87.35 & 53.20 & 66.40 \\
+ LLaVA-Reward-Phi~\cite{zhou2025multimodal} & 79.03 & 77.35 & 78.90 & 85.07 & 48.40 & 65.20 \\
+ AR-GRPO~\cite{yuan2025ar}& 80.55 & 78.89 & 80.55 & 85.88 & 53.20 & 66.56 \\
+ X-Omni~\cite{geng2025x} & 82.07 & 77.79 & 79.44 & 85.53 & 50.80 & 65.24  \\
\rowcolor{gray!20}
+ RubricRL (Ours)  & \textbf{82.98} & 77.93 & 79.74 & \textbf{86.92} & \textbf{54.80} & \textbf{67.70} \\
\bottomrule
\end{tabular}
\label{tab:diffusion_dpg}
\vspace{-1.em}
\end{table}

\subsection{Comparing with State-of-the-arts}

We compare RubricRL with representative reward modeling strategies for RL post-training on two widely used benchmarks, DPG-Bench~\cite{hu2024ella} and GenEval~\cite{wei2024general}, covering both autoregressive (AR) and diffusion-based text-to-image generators. The compared methods can be grouped by reward design: (i) \emph{single} specialized reward models, including CLIPScore~\cite{radford2021learning}, HPSv2~\cite{wu2023human}, Unified Reward~\cite{wang2025unified}, and LLaVA-Reward-Phi~\cite{zhou2025multimodal}; and (ii) \emph{composite} reward metrics with fixed weights, such as AR-GRPO~\cite{yuan2025ar} and X-Omni~\cite{geng2025x}. For fair comparison, we implement these baselines under the same RL framework and hyperparameters as ours, and only vary the reward function. We also report the initial SFT models to isolate the gains brought by RL.

\noindent\textbf{Autoregressive backbones.}
We evaluate the above reward baselines on two AR SFT backbones (Phi3 and Qwen2.5). Quantitative results are reported in Table~\ref{tab:geneval} (GenEval) and Table~\ref{tab:dpgbench} (DPG-Bench). For GenEval, we apply prompt rewriting following~\cite{deng2025emerging} to ensure evaluation consistency. For DPG-Bench, since LLM-based generators are trained with long prompts, we also rewrite prompts for a consistent evaluation interface.
From the results, all RL post-trained methods consistently outperform the SFT baseline, confirming the benefit of reinforcement learning in enhancing image generation quality.
And RubricRL achieves the best performance, surpassing X-Omni by approximately 4\% on GenEval on both LLM backbones, highlighting the effectiveness and generalization of our rubric-based reward.

\noindent\textbf{Diffusion backbones.}
Beyond the AR setting, we further evaluate whether RubricRL generalizes to diffusion-based generators under the same benchmarks. As shown in Table~\ref{tab:diffusion_gen}, RubricRL yields consistent gains for both diffusion backbones, improving over the Qwen-Image baseline by +6.85\% and over the FLUX baseline by +8.48\%. We further report DPG-Bench results in Table~\ref{tab:diffusion_dpg}, where RubricRL brings smaller but still consistent improvements. These results indicate that RubricRL is a simple and effective plug-and-play component for reward-driven optimization across model families.

\subsection{Ablation Study}
In this section, we present ablation studies and additional analyses conducted under the autoregressive (AR) setting. Unless otherwise specified, all experiments are performed on the Phi3 backbone and evaluated on GenEval.

\begin{table}[t]
\caption{Comparison of dynamic rollout-sampling strategies used in GRPO.}
\centering
\scriptsize
\setlength{\tabcolsep}{4.2pt}
\begin{tabular}{lccccccc}
\toprule
Method & Single Obj. & Two Obj. & Counting & Colors & Position & Color Attr. & Overall \\
\midrule
Vanilla & 0.9906 & 0.9217 & 0.6031 & 0.9441 & 0.7975 & 0.7525 & 0.8349 \\
FFKC-1D~\cite{gonzalez1985clustering}   & 1.0000 & 0.9268 & 0.5656 & \textbf{0.9441} & 0.8250 & 0.7500 & 0.8353  \\
DAPO~\cite{yu2025dapo} & 0.9969 & 0.9293 & 0.6125 & 0.9362 & 0.7975 & 0.7275 & 0.8333 \\
\rowcolor{gray!20}
Hybrid (Ours) & \textbf{1.0000} & \textbf{0.9343} & \textbf{0.6125} & 0.9415 & \textbf{0.8275} & \textbf{0.7650} & \textbf{0.8468}\\
\bottomrule
\end{tabular}
\label{tab:geneval_dynamic_sample}
\vspace{-0.5em}
\end{table}

\begin{table}[t]
\caption{Comparison of advantage computation in GRPO, with performance measured on GenEval.}
\vspace{-0.5em}
\centering
\scriptsize
\setlength{\tabcolsep}{4pt}
\begin{tabular}{lccccccc}
\toprule
Method & Single Obj. & Two Obj. & Counting & Colors & Position & Color Attr. & Overall \\
\midrule
RubricRL\_GN & 0.9938 & 0.9268 & \textbf{0.6781} & 0.9362 & 0.7850 & 0.6825 & 0.8337\\
\rowcolor{gray!20}
RubricRL\_LN & \textbf{1.0000} & \textbf{0.9343} & 0.6125 & \textbf{0.9415} & \textbf{0.8275} & \textbf{0.7650} & \textbf{0.8468} \\
\bottomrule
\end{tabular}

\label{tab:geneval_scope_adv}
\vspace{-1.em}
\end{table}

\noindent\textbf{Strategies for dynamic rollout sampling.}
To investigate the impact of different selection strategies used by dynamic rollout sampling, we compare four methods, \textit{i.e.}, RubricRL without dynamic rollout sampling (Vanilla), FFKC-1D~\cite{gonzalez1985clustering}, DAPO~\cite{yu2025dapo}, and our proposed hybrid strategy, and report the results in Table~\ref{tab:geneval_dynamic_sample}. Specifically, FFKC-1D also oversamples more rollouts and then keeps a diverse subset by first selecting a medoid (the rollout with reward closest to the median) and then greedily adding samples that maximize reward distance from already chosen ones. Compared to our hybrid strategy, FFKC-1D focuses too much on the diversity and ignore the importance of high quality rollouts.
As shown in Table~\ref{tab:geneval_dynamic_sample}, our hybrid sampling strategy consistently achieves the best performance, surpassing both FFKC-1D and DAPO as well as the Vanilla baseline that directly uses four rollouts without any dynamics. Interestingly, FFKC-1D and DAPO do not outperform the vanilla baseline, suggesting that their dynamic prompt sampling and pure rollout diversity-driven  sampling strategies fail to provide additional informative signals for RL. In contrast, our hybrid strategy effectively balances exploitation of high-reward rollouts and exploration of diverse candidates, enabling the policy model to leverage both higher-quality and diverse samples, resulting in more effective RL signal.

\noindent\textbf{Normalization scope for advantages.}
In Eq. \ref{eq:adv}, the advantage used in GRPO is computed by normalizing rewards—using the mean and standard deviation—
within a group of rollouts.
Under our dynamic sampling strategy, only $N$ rollouts are retained out of $N'$ candidates. This raises an important design choice: should the normalization statistics (mean and standard deviation) be computed using all $N'$ rollouts or only the retained $N$? We denote these two norm variants as ``RubricRL\_GN" and ``RubricRL\_LN", respectively.
In Table \ref{tab:geneval_scope_adv}, it reflects that ``RubricRL\_LN" yields better performance. This is because normalizing within the retained subset better reflects the actual reward distribution that guides learning, preventing high-variance or low-quality rollouts from distorting the gradient direction.

\noindent\textbf{RubricRL v.s. SFT with Best-of-N sampling.}
We further compare the proposed RubricRL with the SFT model equipped with a Best-of-N sampling strategy during inference($N=8$), which has been observed in prior work \cite{geng2025x} to form an `upper bound' for RL methods in language tasks. 
Specifically, for each prompt in GenEval, we first generate a rubric, then sample 8 rollouts from the SFT model. Each rollout is scored using the rubric-based reward, and the top 4 are selected for evaluation on GenEval. As shown in Table~\ref{tab:geneval_bestofn}, although Best-of-N sampling significantly can achieve higher scores, RubricRL still achieves a notable improvement, exceeding it by over 5\%. This result aligns with the observations in X-Omni~\cite{geng2025x}, reconfirming that reinforcement learning provides a more effective optimization paradigm.

\begin{table}[t]
\caption{Comparison on GenEval of Best-of-N ($N=8$) and our RL training for Phi3-3.8B (P-*) and Qwen2.5-0.5B (Q-*).}
\centering
\scriptsize
\setlength{\tabcolsep}{4pt}
\begin{tabular}{lccccccc}
\toprule
Method & Single Obj. & Two Obj. & Counting & Colors & Position & Color Attr. & Overall \\
\midrule
P-SFT model   & 0.9938 & 0.8939 & 0.4562 & 0.9255 & 0.7200 & 0.5850 & 0.7624 \\
P-Best-of-N & 0.9812 & 0.9091 & 0.6000 & 0.9309 & 0.7450 & 0.5900 & 0.7927 \\
\rowcolor{gray!20}
P-RubricRL &  \textbf{1.0000} & \textbf{0.9343} & \textbf{0.6125} & \textbf{0.9415} & \textbf{0.8275} & \textbf{0.7650} & \textbf{0.8468} \\
\midrule
Q-SFT model   & 0.9625 & 0.5303 & 0.2500 & 0.7314 & 0.3575 & 0.2825 & 0.5239 \\
Q-Best-of-N & 0.9562 & 0.6566 & \textbf{0.3750} & 0.8138 & 0.4100 & 0.3600 & 0.5953 \\
\rowcolor{gray!20}
Q-RubricRL & \textbf{0.9844} & \textbf{0.6616} & 0.2469 & \textbf{0.8378} & \textbf{0.4825} & \textbf{0.3950} & \textbf{0.6014} \\
\bottomrule
\end{tabular}
\label{tab:geneval_bestofn}
\vspace{-1em}
\end{table}

\begin{figure}[ht]
\centering

\begin{minipage}[t]{0.51\textwidth}
    \vspace{0pt}
    \centering
    \includegraphics[width=\linewidth]{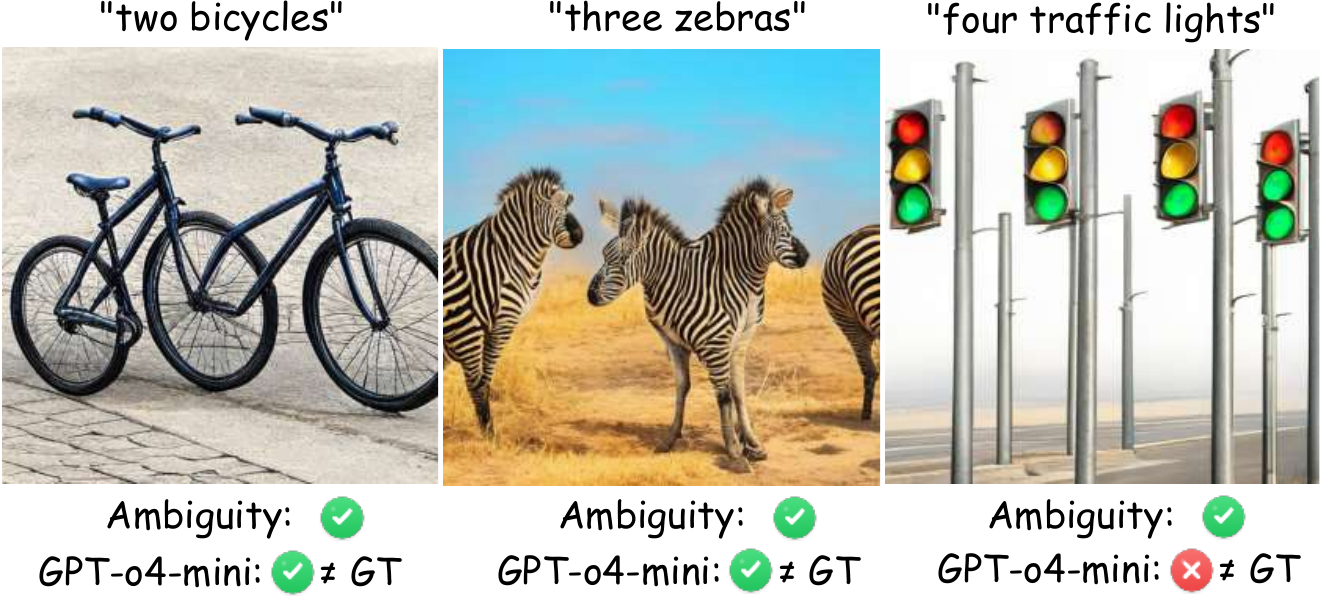}
    \captionsetup{type=figure}
    \vspace{-5pt}
    \caption{Failure cases of GPT-o4-mini when grading counting on GenEval. The model misjudges instance counts under ambiguity. The grader’s score deviates from the GT and thus from common-sense human judgment.}
    \label{fig:o4mini_grader}
\end{minipage}\hfill
\begin{minipage}[t]{0.45\textwidth}
    \vspace{0pt}
    \centering
    \includegraphics[width=\linewidth]{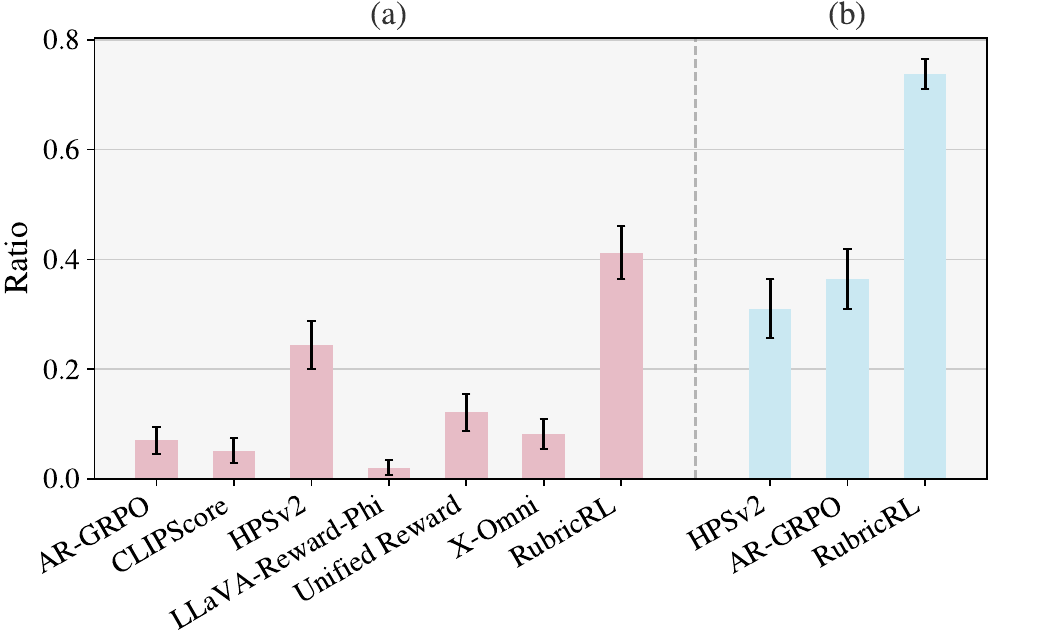}
    \captionsetup{type=figure}
    \vspace{-12pt}
    \caption{Visualization of human preference and judge alignment analysis. (a) Top-1 preference ratio. (b) Human preference alignment ratio.}
    \label{fig:human_pref}
\end{minipage}
\vspace{-1.5em}
\end{figure}

\noindent\textbf{Failure case analysis.} As the grader, although GPT-o4-mini is highly general and powerful in evaluating the quality of generated images, we observe that it can assign incorrect scores—\textit{e.g.}, underestimating or overestimating object counts, particularly when the base model’s generation quality is poor. Fig.~\ref{fig:o4mini_grader} illustrates several typical failure cases in the counting subcategory of GenEval, such as redundant poles near traffic lights, intertwined bicycles, and overlapping zebras. These challenging scenarios often mislead GPT-o4-mini, resulting in inaccurate counts. However, when the base model generates higher-quality images, this issue is less pronounced. This explains why RubricRL’s performance on the “counting” subcategory in GenEval and the “Other” subcategory in DPG-Bench—both containing many counting cases—is worse than the baseline SFT model when using Qwen2.5-0.5B as the base model. In contrast, with Phi3-3.8B, this issue almost disappears, allowing RubricRL to substantially improve performance in counting-related categories.

\noindent\textbf{Rubric reliability and grader correctness.}
To validate our rubric-based training signal, we run a human study on rubric quality and grader reliability. We sample 100 prompts, generate 10 rubrics per prompt, and evaluate 1,000 prompt--rubric pairs with ten annotators for (i) prompt--rubric consistency, (ii) rubric hallucination, and (iii) agreement between GPT-o4-mini and human judgments. We find that 98.1\% of rubrics are consistent with the prompt and only 0.95\% contain hallucinated constraints. Moreover, GPT-o4-mini matches human judgments in 94.8\% of cases, supporting its use as a reliable grader.

\noindent\textbf{Human preference and judge alignment analysis.}
We conduct two user studies: (1) human preference among final models trained with different rewards, and (2) agreement between human and reward-model judgments. For (1), we sample 100 prompts, generate outputs from each method, and ask users to pick the best image. Fig.~\ref{fig:human_pref}(a) shows RubricRL is chosen most often, indicating higher human preference. For (2), we compare RubricRL with HPSv2 (unified) and AR-GRPO (scalar): each method produces two images per prompt, users select the better one, and we compute agreement with the model judge. Fig.~\ref{fig:human_pref}(b) shows RubricRL achieves the highest agreement with lower variance.

\begin{figure}[t]
  \centering
\includegraphics[width=\linewidth]{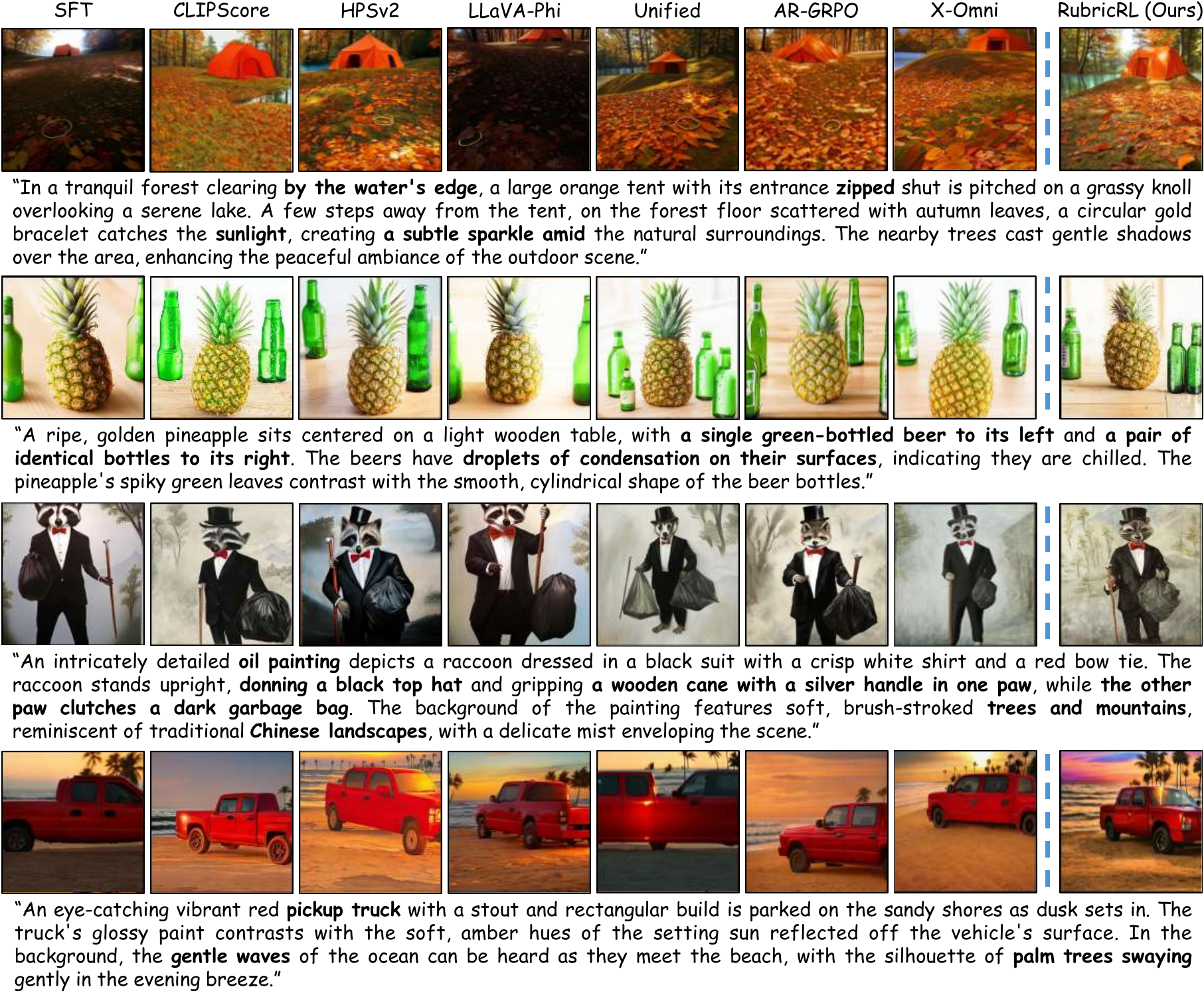}

  \caption{Qualitative comparison: we visualize RubricRL and baseline models using prompts from DPG. RubricRL shows superior image quality that is both aesthetically pleasing and better aligned with the prompt. The \textbf{bold text} highlights key elements that RubricRL successfully captures, while baseline models often fail to generate these details accurately.
}
  \label{fig:sota_visual}
  \vspace{-2em}
\end{figure}

\subsection{Visual results}
We further present comprehensive visual comparisons between RubricRL and other baseline approaches in Fig.~\ref{fig:sota_visual}. As illustrated, models trained with RubricRL consistently produce images that are not only more aesthetically appealing but also demonstrate superior semantic alignment with the given input prompts. To aid interpretation, any misaligned or missing elements in the generated images are emphasized using bold text. 
In particular, LLaVA-Reward-Phi \cite{zhou2025multimodal} and UnifiedReward \cite{wang2025unified} generate outputs where the black bag is not properly held in hand, and in some cases, depict two bags in both paws while omitting the wooden cane altogether. These qualitative observations underscore the effectiveness of RubricRL in enhancing the model’s capability to follow complex, fine-grained instructions and produce high-quality, prompt-consistent images. 
\section{Conclusion}
In this paper, we introduce RubricRL, a rubric-based reward RL framework that provides prompt-adaptive, decomposable supervision for text-to-image generation. By explicitly creating configurable visual criteria (\textit{e.g.}, counting, attributes, OCR fidelity, realism) and scoring them independently, RubricRL produces interpretable and modular signals that integrate seamlessly with standard policy optimization in RL. Experimental results demonstrate that RubricRL outperforms existing RL-based approaches for enhancing text-to-image generation. We hope this work provides new insights into applying reinforcement learning to visual generation models.


%
%
\bibliographystyle{splncs04}
\bibliography{main}

\appendix
\clearpage
\setcounter{page}{1}
\title{Supplementary Material of ``RubricRL: Simple Generalizable Rewards for Text-to-Image Generation''} 

\author{}
\institute{}
\maketitle

\noindent In the following sections, we first introduce the additional methodology details, including the differences between autoregressive and diffusion models, and details about DiffusionNFT as a reinforcement learning framework. Then we provide more ablation results and visualization results to further demonstrate the effectiveness of our method.

\section{More Methodology Details}
\label{sec:diffusionnft}

\subsection{Prompt Templates for Rubric Generation and Evaluation}
\label{sec:prompt_templates}

To improve reproducibility, we provide the three MLLM prompts used in RubricRL. 
The pipeline contains three stages. 
First, the candidate rubric-generation prompt asks the MLLM to identify prompt-relevant criteria from an editable rubric-key pool, covering dimensions such as OCR alignment, object count consistency, attribute accuracy, spatial reasoning, scene coherence, aesthetic quality, and style consistency. 
Second, when the candidate set is large, the rubric-selection prompt selects the most important criteria by prioritizing prompt-grounded fidelity while preserving visual quality and plausibility. 
Third, the image-evaluation prompt scores each generated image against one selected criterion at a time, returning \textbf{1} if the image satisfies the criterion and \textbf{0} otherwise. 
This three-stage prompting design makes the reward both reproducible and interpretable: rubric generation determines what should be evaluated, rubric selection keeps the reward compact, and binary image evaluation measures whether each explicit visual requirement is satisfied.

\subsection{Prompt Templates for Rubric Generation and Evaluation}

To improve reproducibility, we provide the three MLLM prompts used in RubricRL.
The prompting pipeline contains three stages:
(1) generating prompt-specific candidate rubrics from the input text prompt,
(2) selecting the most relevant rubrics when the candidate set is large, and
(3) evaluating each generated image against each selected rubric with a binary score.
The placeholder \texttt{<image>} denotes the generated image input.

\begin{figure}[!ht]
\centering
\begin{minipage}{0.98\linewidth}
\prompttitle{Prompt for Candidate Rubric Generation}
\begin{lstlisting}[style=promptstyle]
You are an expert in evaluating the quality of images generated by text-to-image models. A user will provide a text prompt. Your task is to reason about the prompt and identify the key rubrics that real users would use to evaluate the generated image. You should consider the rubric categories listed below only if they are relevant to the prompt, and you may add additional rubrics based on your expert judgment.

1. OCR Alignment: If the text prompt explicitly or implicitly implies or requires text to appear in the image, evaluate whether any textual content (e.g., labels, signs, clothing text) within the generated image matches the OCR text that is implied or stated in the prompt. If the text prompt says no OCR text/specific OCR text, you should have a corresponding rubric as well. If the given text prompt does not contain any OCR text related requirement, please do not return OCR-related rubric.

2. Object Count Consistency: Include this if the prompt explicitly or implicitly specifies the number of key objects or subjects. Assess whether the number of objects described in the prompt is accurately reflected in the image. If no specific object count is mentioned in the text prompt, please do not return this count-related rubric.

3. Attribute Accuracy: Include this if the prompt explicitly describes any appearance/attributes. Examine whether the attributes (e.g., adjectives like "red", "shiny", "texture") of each object in the image match those specified in the prompt.

4. Action Accuracy: Include this if the prompt explicitly describes any actions/facial expression. Examine whether the actions (e.g., "running", "holding") of each object in the image match those specified in the prompt.

5. Object Placement and Spatial Reasoning: Include this if the prompt implies or specifies spatial relations between objects. Evaluate whether the spatial relationships or positioning of objects in the image correspond to what is expected from the prompt (e.g., "the dog is sitting under the table").

6. Scene Coherence: Assess if the background, lighting, and object composition are visually coherent and contextually appropriate. Assess both the visual appeal of the image and whether the background scene or co-occurring objects are contextually and semantically appropriate given the primary objects described in the prompt.

7. Aesthetic Quality: If the given text prompt does not explicitly require the generated image to be a specific aesthetic level (e.g., blurry, noisy), assess whether the generated image has a high aesthetic quality. Otherwise, evaluate whether the generated image meets the required aesthetic quality.

8. Style and Prompt Consistency: Include this if the prompt mentions visual or artistic style. Assess whether the image adheres to the artistic or visual style stated in the prompt (e.g., "a watercolor painting", "abstract"). Check for stylistic fidelity and absence of unintended visual styles.

Please output up to 10 rubrics in a single JSON object.
Each key should be a short rubric name, and each value should be a brief explanation.

Valid example:
{"Object Count Consistency": "Verify that exactly one cat is shown.", "OCR Alignment": "Ensure visible text matches prompt description."}

Invalid example:
"Object Count Consistency","Verify that exactly one cat is shown."

Strictly follow this format:
- A single JSON object, not a list.
- Keys and values wrapped in double quotes.
- No trailing commas or comments.
- No line breaks within keys or values.
- The output must be valid JSON parsable with Python json.loads().

<EXAMPLE>
Input text prompt: "Generate a photo that a white cat is playing soccer with a brown dog wearing a navy-white stripe sweater."

Output rubrics:
    {"Object Count Consistency (White Cat)": "Verify that exactly one white cat is present in the image, as specified by the prompt.", "Object Count Consistency (Brown Dog)": "Verify that exactly one brown dog is present in the image, as specified by the prompt.", "Attribute Accuracy (White Cat Color)": "Verify that the cat's fur is solid white without stray colors or markings.", "Attribute Accuracy (Dog Sweater)": "Verify that the dog is wearing a sweater with a clearly defined navy-and-white striped pattern.", "Action Accuracy (Soccer Play - Dog)": "Verify that the brown dog is engaged in a soccer-playing action, e.g., kicking or dribbling a soccer ball.", "Scene Coherence": "Verify that the background environment and lighting are contextually appropriate for a soccer game, e.g., grassy field, goalposts, stadium.", "Aesthetic Quality": "Verify that the final image is high quality: sharp focus, realistic lighting and shadows, natural textures, and absence of rendering artifacts."}
</EXAMPLE>
\end{lstlisting}
\end{minipage}
\caption{Prompt template for candidate rubric generation. The MLLM receives the user prompt and an editable rubric-key pool, then outputs prompt-relevant rubrics as a JSON object.}
\label{fig:prompt_rubric_generation}
\end{figure}

\begin{figure}[!ht]
\centering
\begin{minipage}{0.98\linewidth}
\prompttitle{Prompt for Rubric Selection}
\begin{lstlisting}[style=promptstyle]
You are an expert in multimodal image evaluation. Your task is to select the ten most important and relevant evaluation criteria (rubrics) for a given image generation prompt.

You are given:
- A user prompt describing the image generation task.
- A rubric_criteria_dict: a dictionary of candidate rubrics, where each key is a criterion name and the value is its detailed description.

Your goal is to maximize prompt-grounded alignment (GenEval indicators) while ensuring basic visual quality and fidelity (DPG-Bench indicators).
The final selection must strictly follow the priority and slot allocation below:

I. Prompt-Grounded Fidelity (GenEval-priority, 6-7 slots)

You MUST fill at least 6 slots, up to 7, with rubrics that directly test whether the generated image matches the prompt.

Always include, if applicable:
- Object identity and presence: object identity, object presence.
- Attribute match: color, size, shape, color accuracy, attribute match.
- Spatial or positional alignment: position alignment, layout accuracy, spatial logic.
- Counting and quantity reasoning: counting accuracy, object number match, group detection.
- Text and OCR: text legibility, OCR match, font clarity.

If the prompt does not explicitly mention some of these aspects, do not invent rubrics, but fill the remaining GenEval slots with the most relevant object-grounded rubrics available.

II. Visual Quality and Plausibility (DPG-priority, 3-4 slots)

After filling GenEval slots, use 3-4 remaining slots for general visual quality and fidelity.

At least two of the following DPG-critical rubrics MUST always be included, if available in rubric_criteria_dict:
1. Realism or physical consistency: realism, physical consistency.
2. Scene and context consistency: scene consistency, object-context fit, background clarity.

Other optional DPG rubrics to consider:
- style match, artistic coherence.
- emotion expression or gesture correctness, only if explicitly described in the prompt.

III. Rules and Output Format

- Return a selected set of key-value pairs from rubric_criteria_dict as a JSON object.
- Do not invent or modify rubrics.
- Strictly follow the slot allocation:
  - 6-7 GenEval-priority rubrics first.
  - 3-4 DPG rubrics next.
- If more than 10 rubrics are relevant, select only those most critical to evaluating success or failure of the generated image.

Input format:
rubric_criteria_list = {rubric_criteria_list}

Prompt: "{user_prompt}"

Return a JSON list with the most important criteria from rubric_criteria_list.
\end{lstlisting}
\end{minipage}
\caption{Prompt template for rubric selection. When the candidate rubric set is large, the MLLM selects the most relevant criteria by prioritizing prompt-grounded fidelity and visual quality.}
\label{fig:prompt_rubric_selection}
\end{figure}

\begin{figure}[!ht]
\centering
\begin{minipage}{0.98\linewidth}
\setlength{\promptboxwidth}{\linewidth}
\prompttitle{Prompt for Rubric-Based Image Evaluation}
\begin{lstlisting}[style=promptstyle]
Developer message:

You are evaluating whether a generated image satisfies the following visual quality criterion.
You will receive a user prompt that was used to generate the image and a visual quality evaluation criterion.
Your job is to return 1 if the image fully satisfies the criterion, or 0 if it clearly fails.
Do not explain or elaborate.
Only output: 1 or 0. Do not include explanations, just return the value.

User message:

<image>

The prompt of the generated image that needs to be evaluated is:
"{user_prompt}"

Evaluation Criterion ({key}):
{criterion}

Scoring Instructions:
- Return 1 if the image fully satisfies the criterion.
- Return 0 if the image clearly fails.

Output: 1 or 0 only.
\end{lstlisting}
\end{minipage}
\caption{Prompt template for rubric-based image evaluation. The MLLM receives the generated image, the original user prompt, and one selected criterion, then returns a binary score.}
\label{fig:prompt_image_evaluation}
\end{figure}

\subsection{Difference between AR and Diffusion-based Models}
Autoregressive (AR) and diffusion/flow-matching text-to-image generators differ primarily in the representation they generate \emph{before} decoding it into pixels. 
AR generators produce an image as a sequence of discrete visual codes obtained from a vector-quantized tokenizer. 
Conditioned on a prompt $p$, AR decoding samples a code sequence $\mathbf{z}=(z_1,\ldots,z_L)$ by the standard next-token factorization
\begin{equation}
\pi(\mathbf{z}\mid p)=\prod_{t=1}^{L} \pi(z_t \mid z_{<t}, p).
\end{equation}
Once $\mathbf{z}$ is generated, the codes are reshaped back to a latent grid and decoded by the tokenizer decoder to obtain the image.

In contrast, diffusion-based and flow-matching generators operate directly in a continuous latent space.
Instead of predicting discrete codes, they start from a noisy latent and iteratively transport it toward a clean latent representation conditioned on the same prompt $p$. 
Under the velocity parameterization, the model predicts a time-dependent velocity field $v_{\theta}(x_t,p,t)$ that specifies how the latent state $x_t$ should evolve over time; sampling integrates these dynamics across steps to obtain a final latent, which is then decoded into pixels by a continuous decoder (\textit{e.g.}, a VAE decoder).
Therefore, prior to pixel decoding, AR models generate \emph{discrete} visual-token sequences via categorical next-token prediction, whereas diffusion/flow-matching models generate \emph{continuous} latents via iterative latent transport driven by a learned velocity field.

\subsection{Reinforcement Learning with DiffusionNFT}
The above comparison highlights that autoregressive and flow-matching generators differ in their pre-decoding generation mechanisms (discrete token prediction vs.\ continuous latent transport). Nevertheless, RubricRL is agnostic to this choice, since the rubric-based reward is computed on the decoded images and can be used as a black-box training signal for different generator parameterizations.

To demonstrate that RubricRL is not restricted to autoregressive generators, we further apply the same rubric-based reward to fine-tune diffusion-based text-to-image models using Diffusion Negative-aware FineTuning (DiffusionNFT)~\cite{zheng2025diffusionnft}.
DiffusionNFT performs policy optimization in the \emph{velocity} parameterization and adopts \emph{within-prompt} (group-relative) reward normalization to stabilize learning under heterogeneous reward scales.

\paragraph{Optimality probability from rubric rewards.}
For each prompt $p$, we sample a group of $K$ images $\{I^{(k)}\}_{k=1}^{K}$ from a data-collection policy and compute the raw reward for each rollout using our rubric reward:
\begin{equation}
R_k \;=\; R_{\text{rubric}}\!\big(I^{(k)},\,p,\,\mathcal{C}(p)\big),\qquad k=1,\ldots,K.
\end{equation}
DiffusionNFT then converts $R_k$ into a bounded \emph{optimality probability} $\rho_k\in[0,1]$ via group normalization:
\begin{equation}
\rho_k
=\frac{1}{2}+\frac{1}{2}\,\mathrm{clip}
\!\left(
\frac{A_k}{M},\;-1,\;1
\right),
\qquad
A_k = \frac{R_k - \bar{R}_p}{\sigma_p + \epsilon},
\label{eq:diffnft_optprob}
\end{equation}
where $\bar{R}_p=\frac{1}{K}\sum_{k=1}^{K}R_k$ and $\sigma_p$ are the within-prompt mean and standard deviation of rewards over the $K$ rollouts, respectively; $\epsilon>0$ is a small constant for numerical stability, $M>0$ is a clipping threshold (set to $5$ in practice), and $\mathrm{clip}(x,-1,1)$ truncates $x$ to the interval $[-1,1]$.

\paragraph{Velocity-space policy optimization.}
Given a mini-batch of tuples $\{(p,I,\rho)\}$ sampled from the buffer, DiffusionNFT samples a diffusion time $t$ and constructs a noisy state $x_t$ from the clean image $I$ via the forward noising process.
Let $v$ denote the target velocity induced by this forward process, and let $v_{\theta}(x_t,p,t)$ be the current velocity predictor.
To perform off-policy updates, DiffusionNFT forms \emph{implicit positive} and \emph{implicit negative} velocities by mixing the current policy with an old/reference policy $v^{\text{old}}$:
\begin{equation}
\begin{aligned}
v_\theta^{+}(x_t,p,t) &= (1-\beta)\,v^{\text{old}}(x_t,p,t)+\beta\,v_{\theta}(x_t,p,t),\\
v_\theta^{-}(x_t,p,t) &= (1+\beta)\,v^{\text{old}}(x_t,p,t)-\beta\,v_{\theta}(x_t,p,t).
\end{aligned}
\end{equation}
The model parameters are then updated by minimizing a weighted regression objective:
\begin{equation}
\min_{\theta}\;
\rho\,\lVert v_\theta^{+}(x_t,p,t)-v\rVert_2^2
+(1-\rho)\,\lVert v_\theta^{-}(x_t,p,t)-v\rVert_2^2,
\label{eq:diffnft_obj}
\end{equation}
where $\rho$ is the optimality probability defined in Eq.~\eqref{eq:diffnft_optprob}.

\paragraph{Update of the sampling policy.}
After each update, the data-collection policy is refreshed via an exponential moving average (EMA) of the training parameters:
\begin{equation}
\theta_{\mathrm{ema}} \leftarrow \eta_i\,\theta_{\mathrm{ema}} + (1-\eta_i)\,\theta,
\end{equation}
where $i$ is the iteration number, $\theta$ denotes the current training parameters and $\theta_{\mathrm{ema}}$ is used for rollouts.
In our implementation, the decay is capped at $\eta=0.9$ with a short warmup schedule
$\eta_i=\min\!\left(\frac{1+i}{10+i},\,0.9\right)$.
During data collection, we temporarily swap the model weights to $\theta_{\mathrm{ema}}$ for rollouts and restore $\theta$ afterward; the buffer is then refreshed for the next iteration.

Overall, DiffusionNFT offers a simple and stable way to incorporate our rubric-based reward into diffusion policy optimization, echoing the group-relative normalization spirit of GRPO while operating on diffusion velocities.

\section{More Ablations}

\subsection{Continuous vs. Binary Reward Supervision} We compare continuous versus binary reward supervision under the same training setting on GenEval. As shown in Table~\ref{tab:score_type}, continuous rewards achieve 81.70\%, while binary rewards improve performance to 84.68\%. Notably, binary supervision is more effective for discrete correctness criteria with clear pass–fail decisions, such as object counting and multi-object color attribute verification, enabling the model to learn more accurate constraints. In contrast, for perceptually nuanced aspects like color, finer-grained scores can provide denser positive feedback by rewarding incremental improvements even when the output does not cross a binary threshold.

\begin{table}[t]
\centering
\caption{Comparison of score type used in our grader, with performance measured on GenEval.}

\scriptsize
\begin{tabular}{lccccccc}
\toprule
Method & Single Obj. & Two Obj. & Counting & Colors & Position & Color Attr. & Overall \\
\midrule
Continuous    & 0.9969 & 0.9268 & 0.5281 & 0.9601 & 0.8175 & 0.6725 & 0.8170 \\
\rowcolor{gray!20}
Binary (Ours) & 1.0000 & 0.9343 & 0.6125 & 0.9415 & 0.8275 & 0.7650 & 0.8468 \\
\bottomrule
\end{tabular}
\label{tab:score_type}
\end{table}

\begin{table}[t]
\caption{Analysis on different number of selected rubrics using for reward ($M$), with performance measured on GenEval.}
\centering
\scriptsize
\setlength{\tabcolsep}{3pt}
\begin{tabular}{lccccccc}
\toprule
Number of $M$ & Single Obj. & Two Obj. & Counting & Colors & Position & Color Attr. & Overall \\
\midrule
$M=4$ & 0.9906 & 0.9116 & 0.4625 & 0.9176 & 0.7575 & 0.6800 & 0.7866\\
$M=8$ & 1.0000 &\textbf{0.9469} & 0.5469 & \textbf{0.9495} & 0.8175 & 0.7450 & 0.8326\\
\rowcolor{gray!20}
$M=10$ (Ours) & \textbf{1.0000} & 0.9343 & 0.6125 & 0.9415 & \textbf{0.8275} & \textbf{0.7650} & \textbf{0.8468} \\
$M=12$ & 1.0000 & 0.9419 & \textbf{0.6156} & 0.9388 & 0.7950 & 0.7650 & 0.8427 \\

\bottomrule
\end{tabular}
\label{tab:geneval_M}
\end{table}

\subsection{Ablation on the Number of Rubrics $M$}
We conduct an ablation study on the number of rubrics $M$ per prompt and report the results in Table~\ref{tab:geneval_M}. As $M$ increases, the overall performance consistently improves from $M=4$ to $M=10$, indicating that a larger rubric set provides more diverse supervision and a richer reward signal. In particular, increasing 
$M$ leads to clear gains on challenging categories such as Counting, Position, and Color Attribute. However, the improvement saturates once $M$ becomes larger. Although $M=12$  achieves a slightly better result on Counting, its overall score is lower than that of $M=10$, suggesting that excessively large rubric sets may introduce redundant or weakly relevant rubrics and thus increase training noise. Therefore, we choose $M=10$ as a balanced trade-off between rubric diversity and redundancy, achieving the best overall performance. 

\subsection{Analysis of Using Different Models as Graders}
Our method, \textit{i.e.}, RubricRL, benefits from a high-quality grader (GPT-o4-mini) in RL: only when per-criterion judgments (\textit{e.g.}, counting, spatial relations, color) are accurate does the reward become informative enough to drive useful policy updates. A weak or noisy grader produces misaligned signals that the policy can overfit or exploit, thereby hurting stability and sample efficiency. By contrast, a reliable grader yields low-noise, goal-aligned rewards that assign credit to the right behaviours and penalize specific errors, making RubricRL effective.

To quantify this effect, we use different vision language models as the grader in RubricRL and report the results in Table~\ref{tab:geneval_grader}. We choose the Qwen2.5-VL~\cite{bai2025qwen2} family with varying model sizes (3B, 7B, and 32B) to evaluate each rollout during training. We observe that the 32B grader clearly outperforms both the 3B and 7B variants, confirming that a stronger vision--language model provides more informative and reliable rewards overall. The 7B model shows a slight improvement over the 3B model, consistent with its higher capacity, while the 3B grader still offers useful signals on certain criteria (\textit{e.g.}, color and position). Nevertheless, the smaller Qwen2.5-VL graders remain noticeably weaker than the 32B grader while all Qwen2.5-VL graders still lag behind the o4-mini grader with a clear gap, which we attribute to o4-mini's stronger instruction following, better multi-step reasoning, and tighter alignment with our rubric design, resulting in sharper, lower-noise per-criterion rewards and ultimately better downstream generation quality.

\begin{figure}[!ht]
  \centering
  \includegraphics[width=1.0\linewidth]{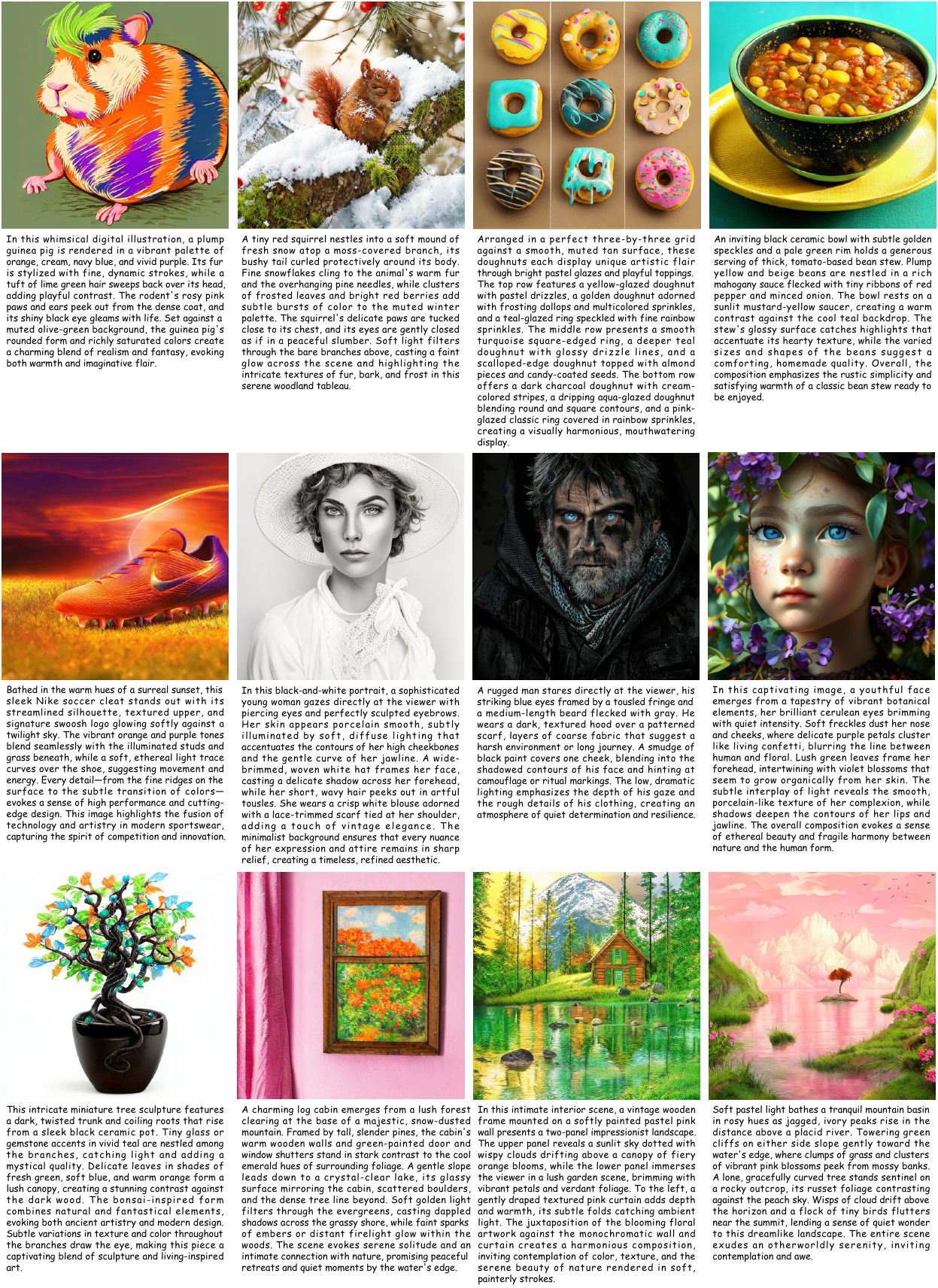}
  \caption{More qualitative results showcasing diverse generations produced by our RubricRL model. The samples exhibit strong prompt following, stylistic versatility, and detailed visual quality.}
  \label{fig:visual}
\end{figure}

\begin{figure}[!ht]
  \centering
  \includegraphics[width=1.0\linewidth]{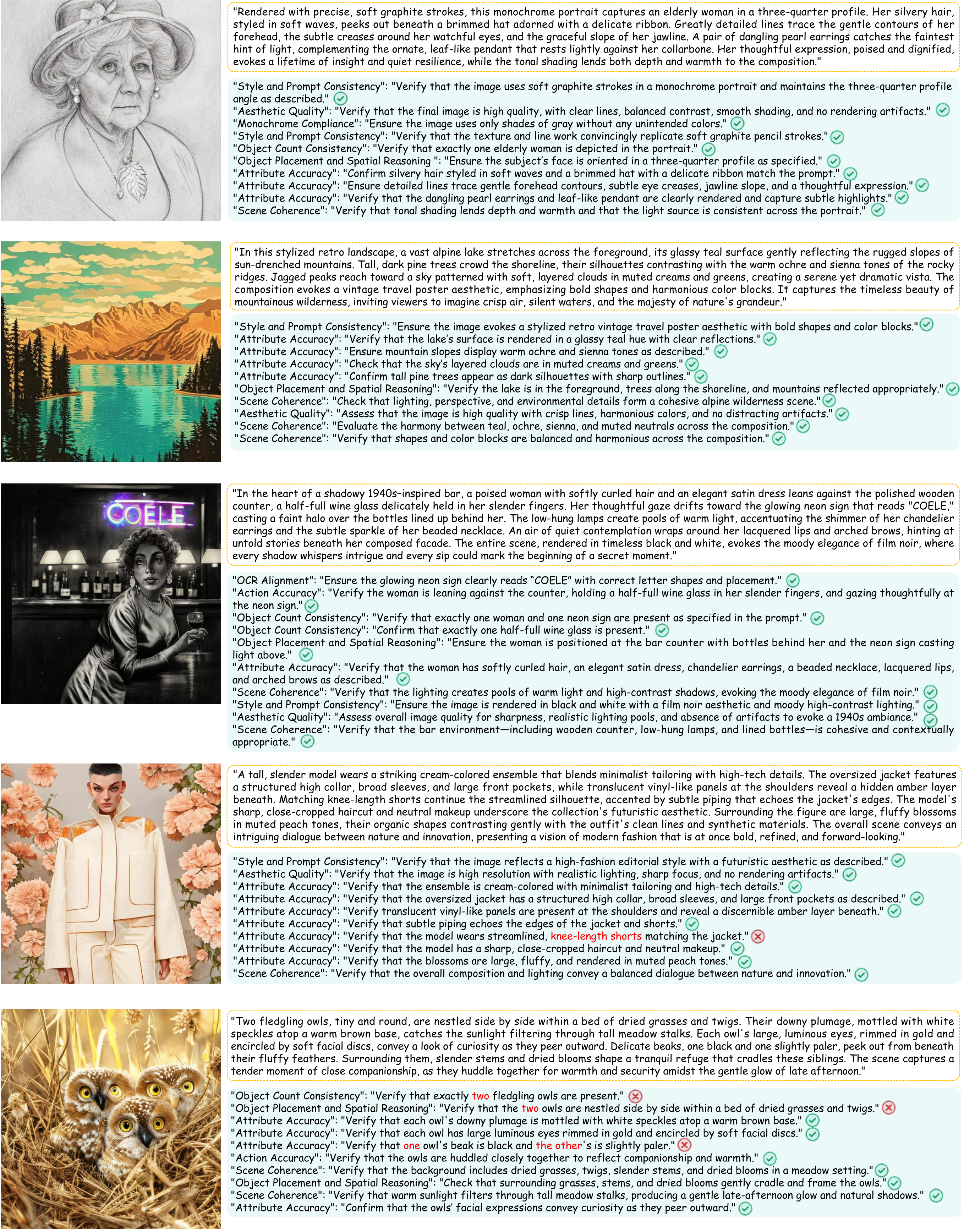}
  \caption{Visualization of our rubric-based reward. For each prompt, we generate evaluation  key–criterion rubrics and score the generated rollout (image) per criteria.}
  \label{fig:visual_rubric}
\end{figure}

\begin{table}[t]
\caption{Comparison of different grader models, with performance measured on GenEval.}
\centering
\scriptsize
\setlength{\tabcolsep}{3pt}
\begin{tabular}{lccccccc}
\toprule
Grader & Single Obj. & Two Obj. & Counting & Colors & Position & Color Attr. & Overall \\
\midrule
Qwen2.5-3B & 0.9938 & 0.9141 & 0.4562 & 0.9441 & 0.7875 & 0.6475 & 0.7906\\
Qwen2.5-7B & 0.9969 & 0.8990 & 0.5031 & 0.9415 & 0.7425 & 0.6925 & 0.7959\\
Qwen2.5-32B & 0.9969 & 0.9268 & 0.5688 & 0.9149 & 0.7725 & 0.6925 & 0.8121\\
\rowcolor{gray!20}
Ours (o4-mini) & \textbf{1.0000} & \textbf{0.9343} & \textbf{0.6125} & \textbf{0.9415} & \textbf{0.8275} & \textbf{0.7650} & \textbf{0.8468} \\
\bottomrule
\end{tabular}
\label{tab:geneval_grader}
\end{table}

\begin{table}[t]
\caption{Comparison of different numbers $N'$ of oversampled rollouts and different numbers $N$ selected. In the main paper, the setting is $N'=16, N=4$.}
\centering
\scriptsize
\setlength{\tabcolsep}{4pt}
\begin{tabular}{lccccccc}
\toprule
Number of $N'$ & Single Obj. & Two Obj. & Counting & Colors & Position & Color Attr. & Overall \\
\midrule
\multicolumn{8}{c}{$N=4$} \\
\midrule
$N'=8$  & 0.9906 & 0.9293 & 0.6000 & 0.9362 & 0.7950 & 0.7200 & 0.8285 \\
\rowcolor{gray!20}
$N'=16$ & \textbf{1.0000} & \textbf{0.9343} & \textbf{0.6125}& 0.9415 & \textbf{0.8275} & 0.7650 & \textbf{0.8468} \\
$N'=32$ & 0.9906 & 0.9318 & 0.5875 & 0.9388 & 0.7950 & 0.7550 & 0.8331 \\
$N'=64$ & 0.9938 & 0.9268 & 0.5844 & 0.9388 & 0.7950 & 0.7650 & 0.8340 \\
\midrule
\multicolumn{8}{c}{$N=8$} \\
\midrule
$N'=16$ & 0.9969 & 0.9192 & 0.5781 & 0.9362 & 0.8025 & 0.7575 & 0.8317 \\
$N'=32$ & 0.9906 & 0.9141 & 0.6094 & \textbf{0.9468} & 0.8225 & \textbf{0.7775} & 0.8435 \\
$N'=64$ & 0.9875 & 0.9343 & 0.6094 & 0.9388 & 0.8150 & 0.7350 & 0.8367 \\
\bottomrule
\end{tabular}
\label{tab:geneval_num_rollout}
\end{table}

\subsection{Analysis of the Number of Rollouts Before and After Dynamic Sampling}
We investigate how the oversampling budget and the post selection budget, \textit{i.e.}, how many rollouts we generate in the dynamic sampling versus how many we keep for reward computation, affect the model's performance.
For each prompt, we first generate \(N'\) candidate rollouts (\(N' > N\)) and then select \(N\) of them using our Hybrid dynamic sampling strategy; the selected \(N\) samples are used to compute the GRPO loss. All other hyperparameters remain fixed across settings.

As shown in Table~\ref{tab:geneval_num_rollout}, increasing the oversampling budget (\textit{e.g.}, $N' \in \{8,16,32,\\64\}$ with fixed $N=4$) initially improves performance by providing a larger candidate pool from which the Hybrid selector can identify high-reward and diverse rollouts. However, the gains soon saturate because larger $N'$ also introduces higher reward variance, making advantage estimates noisier and hindering stable optimization. A similar phenomenon appears when increasing the selection budget from \(N=4\) to \(N=8\): although more selected rollouts increase exploitation, incorporating too many rollouts increases the likelihood of including low-quality generations, amplifying variance in the group-normalized advantage and diluting the learning signal. Notably, configurations with a 4\(\times\) oversampling ratio achieve comparable overall performance, indicating that maintaining this level of oversampling is sufficient for obtaining high-quality candidates. Overall, both oversampling and selection are beneficial only up to a point—beyond that, the added diversity is outweighed by increased noise, revealing an inherent trade-off between exploration and optimization stability in GRPO-style training.

\subsection{Evaluation on Aesthetic Quality}
The main paper focuses on compositional benchmarks such as GenEval and DPG-Bench, since they directly measure instruction following and compositional correctness, which are the primary targets of RubricRL. However, an important question is whether improving rubric-based alignment comes at the cost of perceptual quality, or whether the learned policy can also benefit broader stylistic and aesthetic aspects of image generation. To address this, we conduct an additional quantitative evaluation using a pretrained CLIP+MLP aesthetic reward model~\cite{christophschuhmann2022improved-aesthetic-predictor} on GenEval. This metric is independent from the rubric rewards used during training and serves as an external proxy for perceptual aesthetic quality. Table~\ref{tab:aesthetic} reports the average aesthetic scores of different reward-learning baselines. As shown in Table~\ref{tab:aesthetic}, RubricRL improves the aesthetic score over the SFT baseline on both model backbones. For Phi3, RubricRL achieves the best score, improving from 4.9621 to 5.3562 and outperforming all compared reward baselines, including CLIPScore, HPSv2, Unified Reward, LLaVA-Reward-Phi, AR-GRPO, and X-Omni. For Qwen-Image, RubricRL also substantially improves the SFT baseline from 5.3297 to 5.6801, achieving the second-best result and remaining competitive with HPSv2, a reward model explicitly designed to capture human preference and aesthetic quality. These results suggest that RubricRL does not merely optimize compositional correctness on standard benchmarks. Instead, rubric-guided learning can also improve perceptual image quality, likely because decomposed rubric feedback provides denser and more stable supervision than a single scalar reward. Therefore, the gains of RubricRL are not limited to compositional metrics, and the method can improve broader visual quality aspects of generated images.

\begin{table}[ht]
\centering
\caption{Additional aesthetic quality evaluation on GenEval using a pretrained CLIP+MLP aesthetic reward model.}

\scriptsize
\begin{tabular}{lclc}
\toprule
& Phi3 (3.8B) & Qwen-Image (4B) \\
\midrule
SFT & 4.9621 & 5.3297 \\
+ CLIPScore & 5.1636 & 5.3085 \\
+ HPSv2 & 5.2860 & \textbf{5.7760}  \\
+ Unified Reward & 5.1769 & 5.5169 \\
+ LLaVA-Reward-Phi  & 4.9580 & 5.3107 \\
+ AR-GRPO & 5.2660 & 5.5747  \\
+ X-Omni & 5.1990 & 5.4653 \\
\rowcolor{gray!20}
+ RubricRL (Ours) & \textbf{5.3562} & 5.6801 \\
\bottomrule
\end{tabular}
\label{tab:aesthetic}
\end{table}

\begin{figure}[ht]
  \centering
  \includegraphics[width=1.0\linewidth]{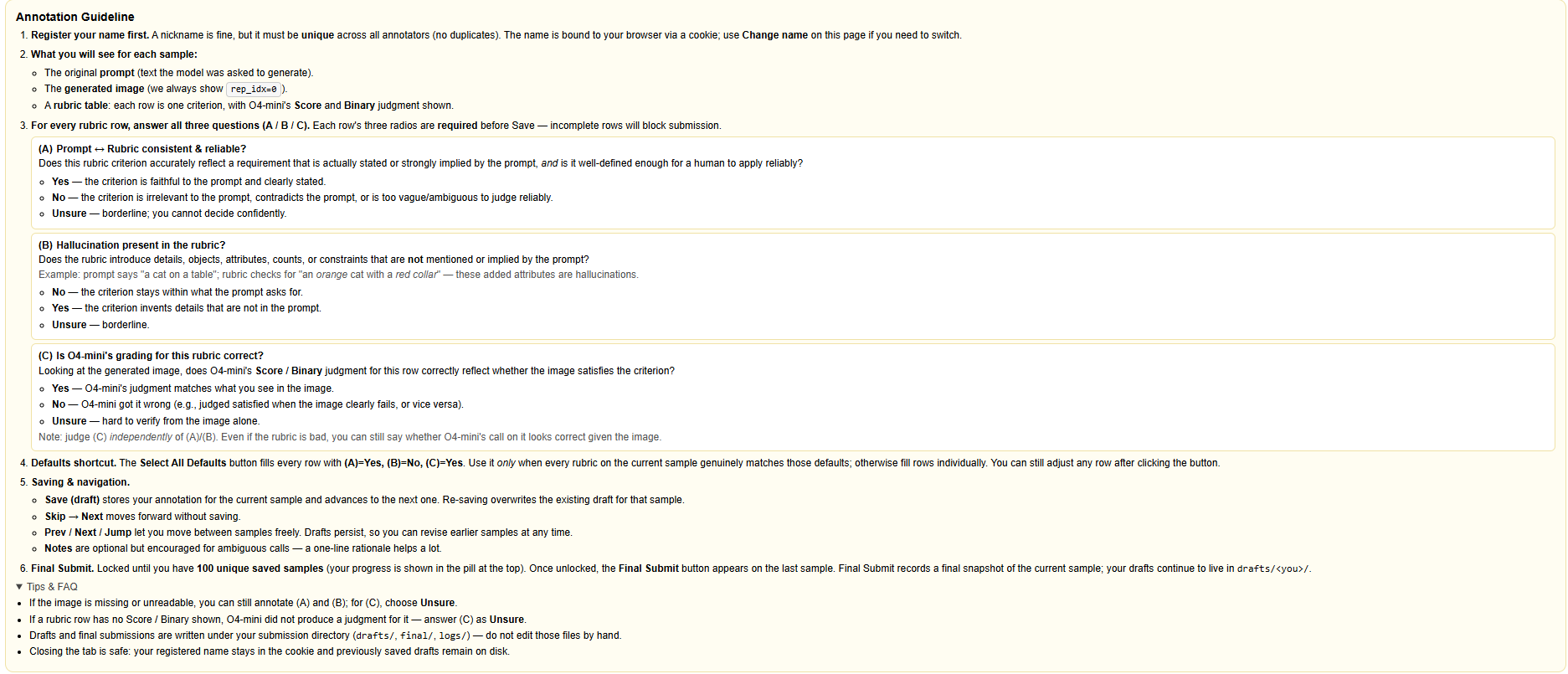}
  \caption{Criteria of rubric reliability and grader correctness user study.}
  \label{fig:reliable}
  \vspace{-1.5em}
\end{figure}

\begin{figure}[ht]
  \centering
  \includegraphics[width=1.0\linewidth]{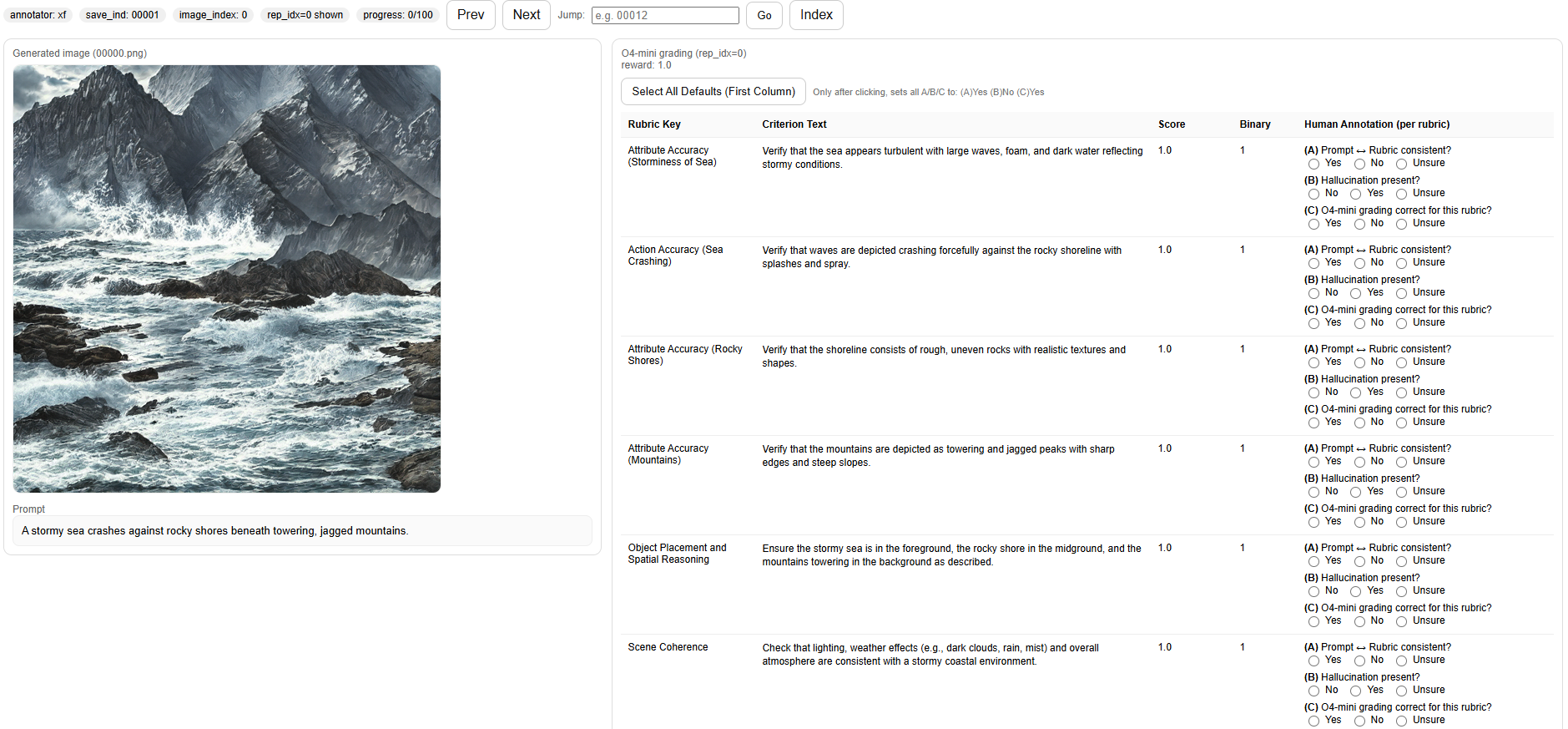}
  \caption{Interface of rubric reliability and grader correctness user study.}
  \label{fig:reliable_interface}
\end{figure}

\begin{figure}[ht]
  \centering
  \includegraphics[width=1.0\linewidth]{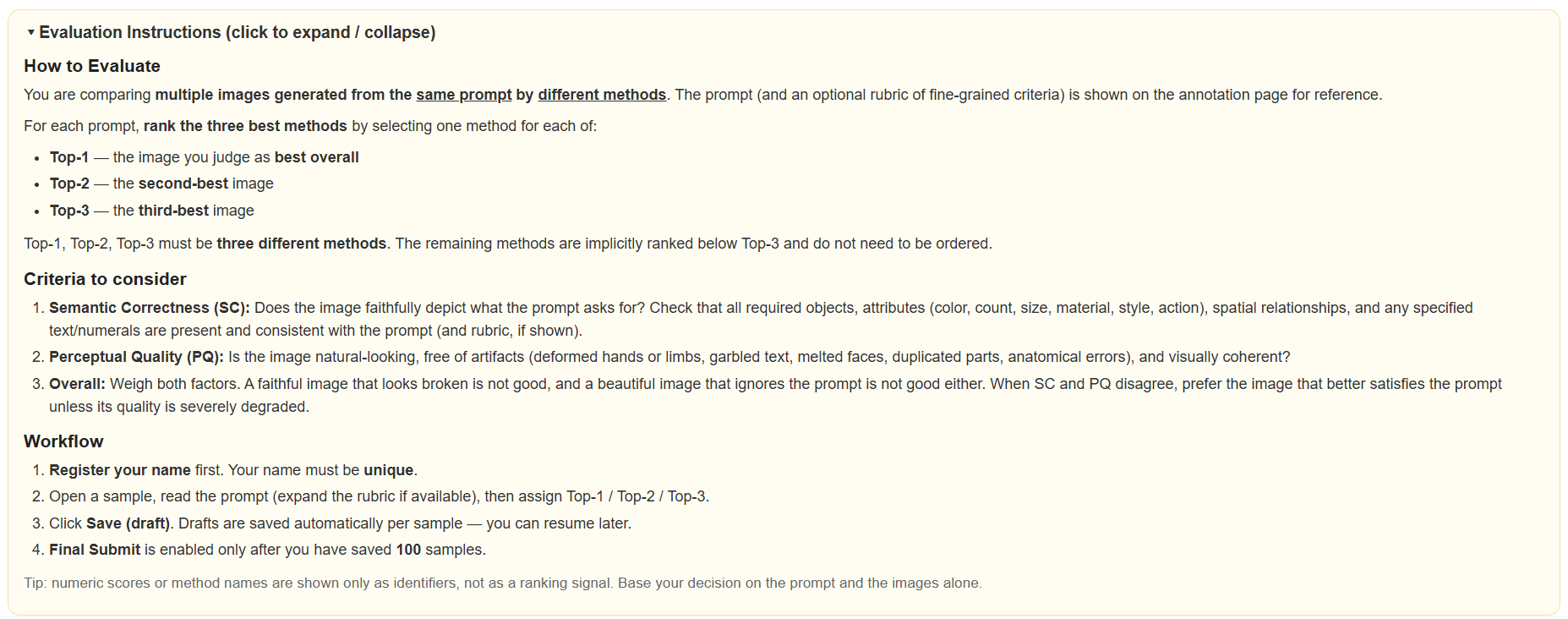}
  \caption{Criteria of top-rank user study. Annotators evaluate generated images based on both prompt and rubrics following and perceptual quality. Images are anonymized and randomly ordered to reduce bias.}
  \label{fig:top_rank}
  \vspace{-1.5em}
\end{figure}

\begin{figure}[ht]
  \centering
  \includegraphics[width=1.0\linewidth]{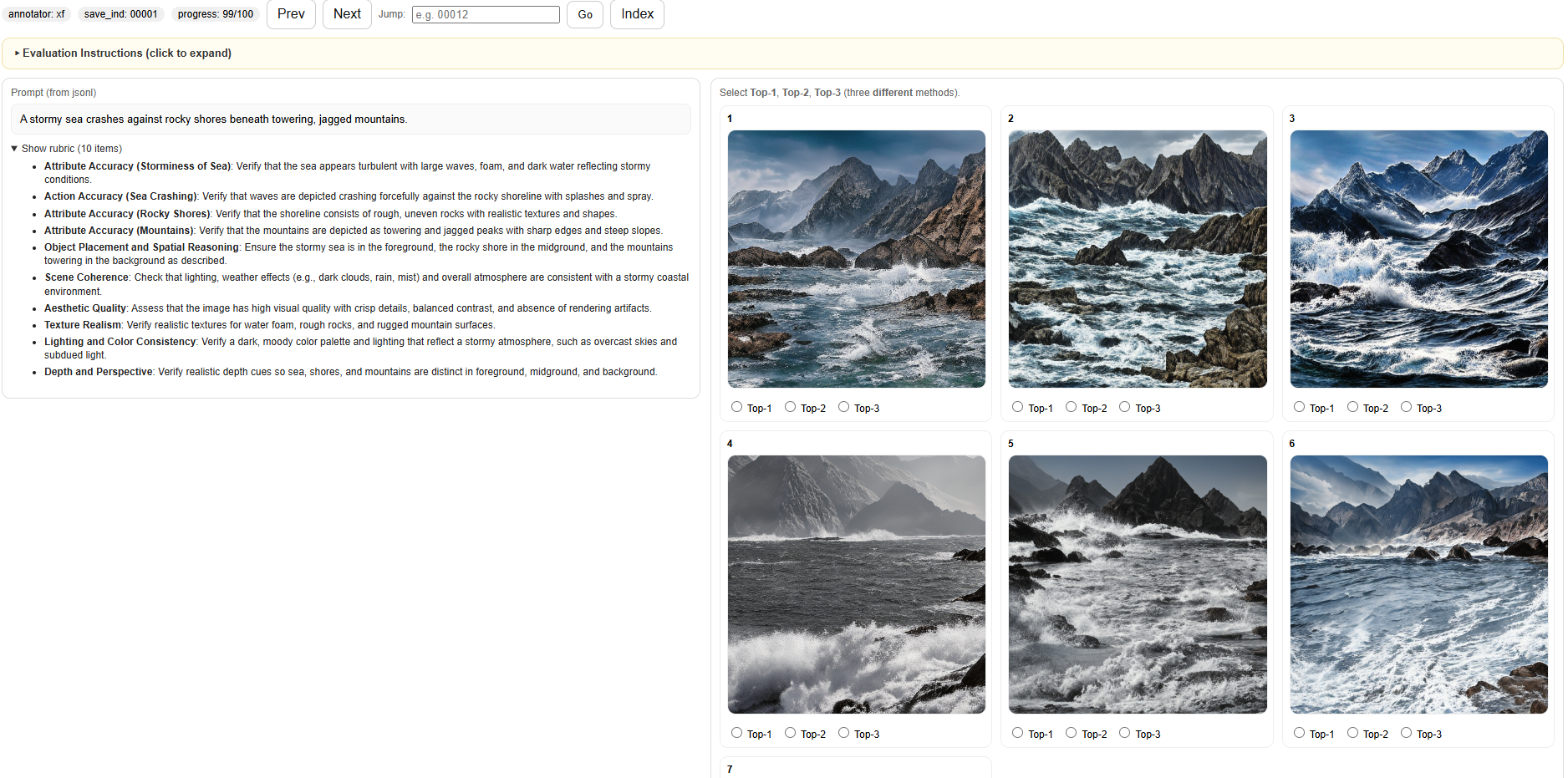}
  \caption{Interface of top1-3 user study. }
  \label{fig:top_rank_interface}
\end{figure}

\begin{figure}[ht]
  \centering
  \includegraphics[width=1.0\linewidth]{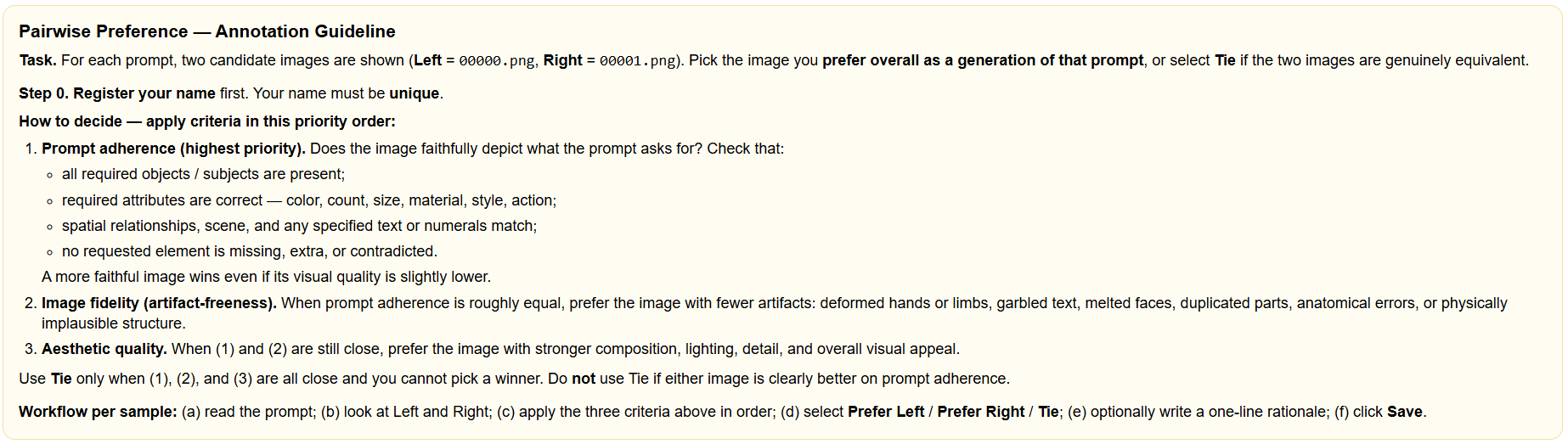}
  \caption{Criteria of pairwise user study. Annotators evaluate generated images based on both prompt following and perceptual quality. Images are anonymized and randomly ordered to reduce bias.}
  \label{fig:pairwise}
\end{figure}

\begin{figure}[ht]
  \centering
  \includegraphics[width=1.0\linewidth]{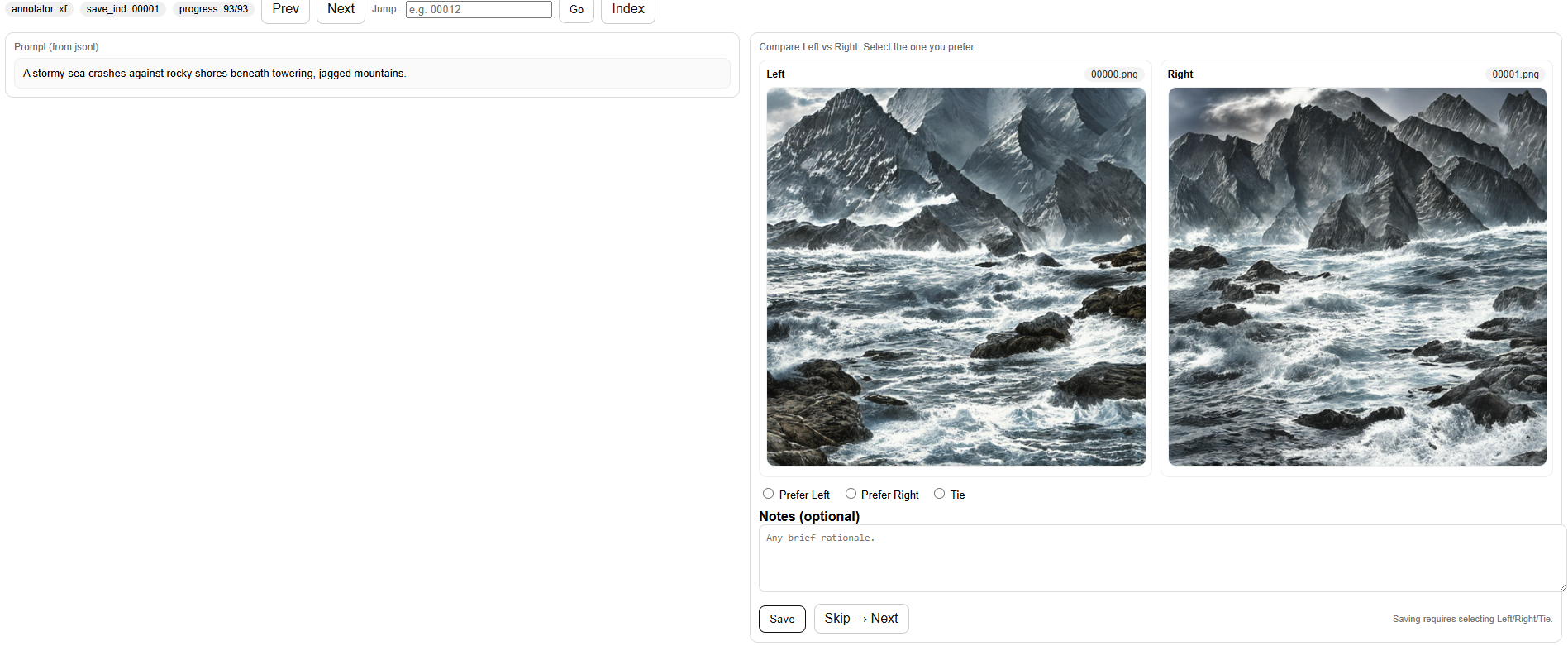}
  \caption{Interface of pairwise user study. }
  \label{fig:pairwise_interface}
\end{figure}

\section{User study Criteria and Interface}
\paragraph{Human study for rubric reliability and grader correctness.}
We provide additional details on the human evaluation used to verify the reliability of our rubric-based supervision. This study is designed to examine whether the automatically generated rubrics are valid with respect to the input prompt and whether the GPT-o4-mini grader produces judgments consistent with human annotators. Fig.~\ref{fig:reliable} shows the annotation instructions and evaluation criteria for this study, and Fig.~\ref{fig:reliable_interface} shows the corresponding annotation interface.

Specifically, we sample 100 prompts and generate 10 rubrics for each prompt, resulting in 1,000 prompt--rubric pairs. Each pair is evaluated by human annotators according to three criteria. First, annotators judge \emph{prompt--rubric consistency}, namely whether the rubric describes a requirement that is explicitly stated or reasonably implied by the prompt. Second, they judge \emph{rubric hallucination}, namely whether the rubric introduces unsupported constraints, objects, attributes, relations, or styles that are not present in the prompt. Third, for grader correctness, annotators compare the binary decision made by GPT-o4-mini with their own judgment for the same criterion and image, and mark whether the two judgments agree.

Across this study, 98.1\% of the generated rubrics are judged to be consistent with the corresponding prompts, while only 0.95\% contain hallucinated constraints. In addition, GPT-o4-mini agrees with human judgments in 94.8\% of the evaluated cases. These results indicate that the generated rubrics provide reliable criterion-level supervision and that GPT-o4-mini can serve as a reliable grader for our training pipeline.

\paragraph{Human preference study criteria and interface.}
We also provide the annotation details for the human preference studies reported in Sec.~4.3 of the main paper. These studies are used to evaluate whether RubricRL improves the final generation quality as perceived by human users. For each evaluation instance, annotators are shown the text prompt together with anonymized generated images. The image order is randomized to avoid positional bias, and method names are hidden from the annotators.

Annotators are instructed to consider two major aspects when making their decisions: \emph{prompt following} and \emph{perceptual quality}. Prompt following includes whether the generated image correctly reflects the requested objects, attributes, counts, spatial relations, actions, and text rendering when such requirements are specified by the prompt. Perceptual quality includes realism, aesthetic appeal, visual coherence, natural composition, and the absence of noticeable artifacts.

We use two complementary annotation protocols. In the top-rank protocol, annotators are asked to select the best image among all candidate methods for the same prompt. This protocol measures which method is most frequently preferred when all results are compared jointly. In the pairwise protocol, annotators compare two images at a time and choose the preferred result. An additional ``Equal'' option is provided when the two images are comparable in both prompt following and perceptual quality. Fig.~\ref{fig:top_rank} and Fig.~\ref{fig:pairwise} present the detailed annotation instructions and evaluation criteria shown to annotators. Fig.~\ref{fig:top_rank_interface} and Fig.~\ref{fig:pairwise_interface} show the corresponding annotation interfaces used in our user study.

\section{Cost, latency, and accessibility.}
RubricRL uses the multimodal grader only during RL training and therefore introduces no inference-time overhead. Before RL training, we generate rubrics offline and cache them as a one-time preprocessing step. During training, for each prompt, we evaluate 16 rollout images against 10 cached rubric criteria. In terms of wall-clock training time, RubricRL remains comparable to multi-reward RL baselines: training RubricRL on Phi3 for one epoch takes about 35 hours, while AR-GRPO with a local multi-reward pipeline takes about 36 hours. For reference, training with CLIPScore alone takes about 17 hours.

Under our usage setting, calling the o4-mini API costs about \$1.18K per epoch according to Azure OpenAI pricing. As a local-compute comparison, the multi-reward pipeline used by AR-GRPO requires 4 A100 GPUs, which costs about \$720 for a 36-hour run under a \$5/A100-hour assumption. Although RubricRL introduces additional training cost, this cost is traded for more fine-grained, controllable, and interpretable supervision. It also improves training efficiency in terms of convergence: on Phi3, AR-GRPO reaches 0.8001 on GenEval after 3 epochs, whereas RubricRL reaches 0.8228 after only 1 epoch. We agree that distilling the judge is a promising direction for reducing cost, latency, and accessibility barriers. However, general-purpose rubric grading is challenging for smaller models because the criteria can be diverse, abstract, and image-dependent. A practical direction is to distill domain-specific graders for narrower rubric categories, or to replace o4-mini with stronger open-source MLLMs as they become available.

\section{Visualization}
In this section, we present additional generations from our RubricRL. As shown in Figure~\ref{fig:visual}, our RubricRL produces high-fidelity images, and significantly improves the model’s ability to follow complex prompts. Additionally, we visualize the detailed key-criterion rubrics for each prompt, along with the correctness or incorrectness of each rollout under each criterion, as shown in Figure~\ref{fig:visual_rubric}.

\end{document}